# Bounded Autonomy and Verifiable Safety for Agentic AI Enabled Automation

Srini Ramaswamy[1], Ph.D.
DNRS.ai, *Columbus OH* USA
srini@computer.org

Deveeshree Nayak
University of Washington, Tacoma, USA
email: dnayak@uw.edu

***Abstract*** - Agentic AI-enabled automation cannot be safely deployed in high-stakes environments on probabilistic reasoning alone. A recurring risk is epistemic drift: as reasoning deepens, system behavior may move away from subject-matter-expert constraints for safe operation. This paper presents BRaVeS, a bounded reasoning and safety-governance framework termed the Defensible Next-Gen Reasoning System (DNRS). BRaVeS encodes SME-defined constraints as invariant anchors, proposes MoDA-Style (Mixture of Depths Attention) depth-aware access as a candidate mechanism for keeping these anchors visible during inference, and uses a state hierarchy (SMARtAutonomy) to reduce autonomy as epistemic risk increases. To formalize bounded recovery, we introduce the Lyapunov-Bounded Consensus Framework (LBCF), which maps continuous epistemic-risk signals into a finite K-bag abstraction and applies shielded state transitions that enforce Lyapunov-style energy descent or route the system to a human-mediated terminal state. The formal convergence result applies to the finite LBCF abstraction under fixed thresholds and feasible-shield assumptions; it does not prove safety of the full continuous neural activation space. We evaluate the framework through a discrete event Monte Carlo simulation using HAI 22.04 industrial-control-system time-series data with synthetic noise and sensor-degradation regimes. Across the tested parameter-grouping strategies and thresholds, the LBCF process achieved finite-step convergence and no safety-guard violations. These results provide simulation-based evidence that bounded governance behavior can be enforced under the stated abstraction, while motivating future work on deployed transformer implementations, live human-in-the-loop validation, and broader adversarial settings.

## I. INTRODUCTION

Agentic AI-enabled automation2 and intelligent autonomous systems cannot enter safety-critical environments on probabilistic reasoning alone. In energy grids, water systems, high-acuity healthcare, defense, and other mission-critical settings, latent-space drift is not merely a model-quality problem; it can become a governance and control problem. Current architectures optimize semantic proximity and task performance, but they do not reliably preserve domain constraints as reasoning deepens. As network depth and agentic delegation increase, SME-defined constraints may become less visible to later reasoning stages, creating vulnerability to epistemic divergence, unsupported action selection, and uncontrolled autonomy. This paper introduces the Defensible Next-Gen Reasoning System (DNRS), a bounded governance architecture that treats generative reasoning as a monitored trajectory rather than an unconstrained probabilistic process. DNRS links probabilistic reasoning to explicit SME-derived boundary conditions, so that agentic AI remains governed by domain-authored constraints rather than unchecked latent optimization. These constraints are encoded as invariant anchors $I(X_0)$, meaning that they are fixed for the governed reasoning episode unless explicitly modified through an authorized verification and validation process. They are not globally immutable rules, nor are they assumed to be permanently internalized by the base model. To support this architecture, the paper makes six contributions. *(i)* First, it defines a method for representing SME rules as structured constraint objects and encoding them into episode-fixed invariant anchors $I(X_0)$. *(ii)* Second, it proposes a MoDA-style invariant-access pathway as a candidate mechanism for keeping SME-derived anchor representations available to later reasoning layers and for supporting divergence estimation, $D = \| I(X_l) - I(X_0) \|$. *(iii)* Third, it introduces the Verification and Validation (V&V) loop, which surfaces contested latent parameters, constructs validator-facing Dossiers, and, where authorized, updates the active invariant reference from $I(X_0)$ to $I(X_0)'$ without retraining the base model. *(iv)* Fourth, it defines the Fuzzy SMARtAutonomy Gearbox, a four-state governance hierarchy - Stable $S$, Meta-Cognitive $M$, Assisted $A$, and Regulated $Rt$ - that reduces operational authority as epistemic risk increases; principally SMARtAutonomy argues that autonomy should not be governed as a binary permission. *(v)* Fifth, it formalizes the Lyapunov-Bounded Consensus Framework (LBCF), which maps continuous epistemic-risk signals into a finite K-bag abstraction and applies shielded transitions requiring either Lyapunov-style energy

[1] Srini Ramaswamy (Corresponding Author)

[2] We use "Agentic AI systems" as shorthand for Agentic AI-enabled automation and intelligent autonomous systems.

descent or Assisted/Regulated terminal routing. *(vi)* Sixth, it evaluates the finite governance abstraction through a 120,000-iteration discrete-event Monte Carlo simulation using HAI 22.04 industrial-control-system time-series data with synthetic

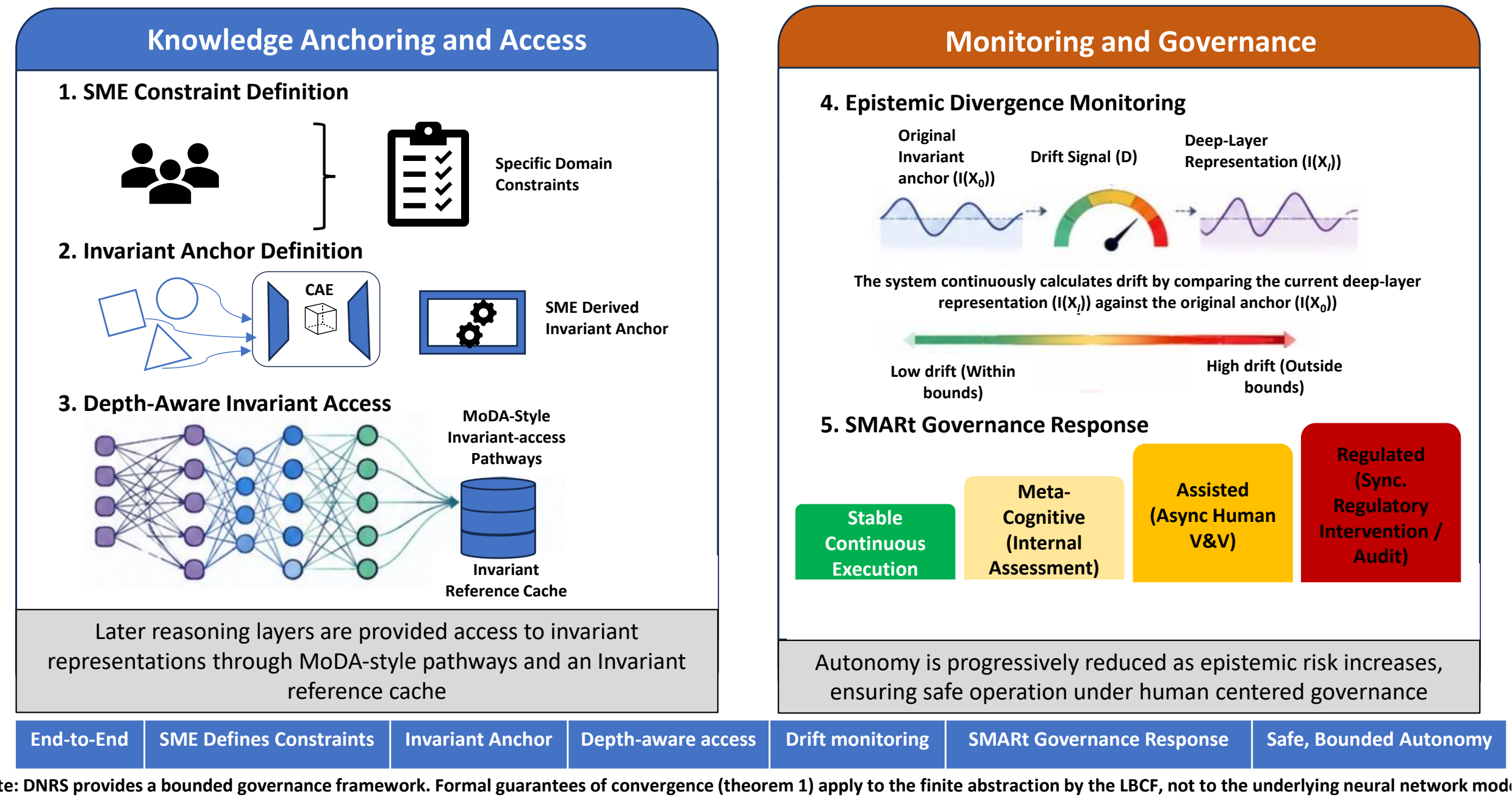

**Figure 1. DNRS-bounded reasoning and safety-governance architecture:** A Step-by-Step process of anchoring AI reasoning to human-defined constraints and dynamically managing autonomy through epistemic monitoring. The framework begins with SME-defined domain constraints, encodes them into an SME-derived invariant anchor $\boldsymbol{I(X_0)}$, provides later reasoning layers with MoDA-style access to invariant representations, monitors epistemic divergence through $\boldsymbol{D} = \| \boldsymbol{I(X_l)} - \boldsymbol{I(X_0)} \|$, and maps increasing epistemic risk into SMARt governance responses. The diagram is architectural rather than implementation-complete: the formal convergence guarantee applies to the finite LBCF abstraction, not to the full continuous neural state.

perturbation regimes.

The paper's claim is intentionally bounded. DNRS is not an AGI blueprint, and it is not designed for unconstrained, domain-independent, or purely exploratory learning. LBCF requires prior SME constraints to define the governed invariant reference $I(X_0)$. MoDA-style invariant access is treated as a proposed architectural mechanism, not as a fully validated production transformer implementation. Direct KV-cache or attention-level access would require model-level access or an equivalent adapter layer; black-box LLM deployments may instead enforce the same governance logic through retrieval gating, tool routing, action suppression, or external policy guards. This paper therefore separates three layers: architecture, formal abstraction, and simulation. DNRS and MoDA-style invariant access provide the architectural proposal for preserving SME-derived constraint visibility. The formal result applies only to the finite LBCF governance abstraction induced by measurable divergence, KL-budget, and epistemic-risk signals. The HAI-based study evaluates whether that discrete governance layer obeys its own convergence and guard-violation rules under synthetic noise, fixed thresholds, and finite K-bag assumptions. It does not claim live deployment, attack-detection accuracy, production MoDA validation, or formal safety of the full continuous neural activation process. To clarify the problem-to-solution structure, Figure 1 summarizes the DNRS pipeline. The architecture separates knowledge anchoring from governance response. SME constraints first define the domain boundary conditions. These constraints are encoded into an invariant anchor $I(X_0)$, which later reasoning layers access through a MoDA-style invariant-access pathway. The system then estimates drift between the active reasoning representation $I(X_l)$and the original anchor. This drift signal becomes the input to SMARt governance and the LBCF finite-abstraction layer, where autonomy is reduced, escalated, or revoked under explicit assumptions. Accordingly, throughout this paper, "verifiable safety" refers to verification of bounded governance behavior over the finite SMARt/LBCF abstraction under stated assumptions, not to a proof of intrinsic safety for the underlying neural model or all possible deployment contexts.

In Section II, we review related work. Sections III through VI detail the core DNRS architecture; including MoDA-style invariant access, the V&V loop, the Fuzzy SMARt gearbox, and the LBCF, qualitatively illustrated through a clinical IDSS workflow. Section VII evaluates bounded-governance behavior through an ICS Monte Carlo simulation under fixed invariant thresholds and finite abstraction assumptions. Section VIII concludes the paper.

## II. BACKGROUND AND MOTIVATIONS

DNRS builds upon four related streams of work; these include: *(i)* safety-control methods, *(ii)* LLM alignment and agentic safety wrappers, *(iii)* architectural mechanisms for invariant access, and *(iv)* state-based autonomy governance.

### A. Directly Comparable Safety-Control Methods

Safe reinforcement learning and constrained decision-making provide the closest technical foundation for the Lyapunov-Bounded Consensus Framework. Foundational Constrained Markov Decision Processes (CMDPs) define policies subject to cost or safety constraints [1]-[6]. Related work on Lyapunov control, time-bounded reachability, control barrier functions, and probabilistic shielding [7] provides mechanisms for enforcing descent, constraint satisfaction, and safe action selection [8]-[14]. These methods are typically applied to physical-control or external decision environments, where the state space represents plant dynamics, actuator choices, or task-level policies. DNRS adapts this safety-control perspective to the governance of internal epistemic conflict: continuous risk signals are not treated as fully verified neural states, but are mapped into a finite K-bag abstraction over which shielded state transitions can be enforced. Under this abstraction, LBCF treats epistemic uncertainty as an energy-like quantity that must decrease, or else trigger Assisted or Regulated intervention.

### B. LLM Alignment and Agentic Safety Wrappers

Modern agentic systems rely on increasingly deep Transformer architectures whose performance is enabled by scaling laws, residual pathways, and depth-dependent representation learning [15] - [18]. However, repeated residual updates can make shallow constraint information less visible as inference proceeds, creating a possible pathway for epistemic drift [19]. Alignment techniques such as RLHF, Constitutional AI, iterative self-reflection, self-critique, and multi-agent debate attempt to improve model behavior through training objectives, critique loops, or external reasoning structures [20]-[25]. These approaches are important, but they do not by themselves specify a finite-state autonomy-governance process tied to measurable invariant drift. DNRS addresses this gap by treating hallucination and unsupported autonomous continuation not only as output-quality failures, but as failures of governance: when epistemic justification weakens, the system must reduce, suspend, or revoke autonomy rather than continue probabilistic optimization.

### C. Architectural Inspiration for Invariant Access

The proposed invariant-access layer draws from neural scaling, sparse routing, and depth-aware model architectures. Early dense cross-layer mechanisms improved access to prior representations but introduced scaling costs that become impractical for very deep systems [26]. More recent data-dependent routing approaches, including Mixture-of-Depths and related MoDA-style mechanisms, provide an architectural inspiration for selectively allocating compute and preserving access to earlier representations [27] - [30]. DNRS uses this idea as a candidate mechanism for keeping SME-derived invariant anchors visible during inference. We do not claim MoDA as a fully implemented or validated transformer architecture and use it as an invariant-access pattern that would require model-level access to internal attention states, KV-cache representations, or an equivalent adapter layer. Therefore, the formal LBCF guarantee does not depend on proving perfect invariant preservation across all hidden layers. It depends on measurable divergence signals being mapped into the finite governance abstraction. Mechanistic interpretability and uncertainty quantification also inform this design. Standard uncertainty quantification methods, including moment propagation, can track mean vectors and covariance structures but may impose substantial computational cost at inference time [31] - [34]. Sparse Autoencoders and related interpretability methods can isolate meaningful features, but their runtime cost and high-dimensional representations may be difficult to integrate into supervisory governance loops. DNRS therefore uses a lower-dimensional abstraction: macro-level residual divergence is evaluated against the SME-derived anchor $I(X_0)$, and epistemic conflict is compressed into ordinal K-bag states. This does not replace full interpretability, but it provides a tractable governance layer for deciding when autonomy should continue, pause, escalate, or halt.

### D. Governance and State-Based Autonomy

DNRS also draws from classical systems verification, computer organization, and state-based agent governance. In hardware and safety-critical software systems, stability is often enforced through architectural constraints, formal guards, and explicit state transitions rather than through post-hoc behavioral correction [35]-[41]. Related “Sensible Agents” and formal state-machine approaches use Extended StateCharts, Petri nets, and local-systemic trade-off models to govern multi-agent behavior [42]-[48]. This tradition motivates the DNRS treatment of SME constraints as explicit invariant anchors rather than as malleable model preferences. Recent work on SMARtAutonomy extends this idea to agentic AI by framing safe autonomy as a state-based lifecycle: systems may operate in Stable mode when constraints are established, shift into Meta-Cognitive mode when internal resolution is possible, enter Assisted mode when SME validation is required, and move to Regulated mode when autonomous action lacks sufficient justification [49]. DNRS builds on this lifecycle by adding invariant anchoring, MoDA-style access, V&V parameter surfacing, and LBCF finite-abstraction control. In this framing, the contribution is not a proof of intrinsic neural safety. It is a bounded governance architecture that links measurable epistemic drift to explicit autonomy-state transitions under stated assumptions.

Together, these bodies of work motivate the structure of DNRS. Safe RL, CMDPs, Lyapunov control, and shielding provide the closest formal basis for bounded state transitions, but they do not directly address epistemic drift inside agentic reasoning systems. LLM alignment and critique methods improve behavior, but they do not by themselves define an autonomy-state lifecycle tied to measurable invariant divergence. Depth-aware and sparse-routing architectures suggest a possible path for preserving access to earlier constraint representations, while state-machine and SMARtAutonomy models provide the governance semantics for suspending, assisting, or revoking autonomy. The following sections combine these threads: Section

III defines SME-derived invariant anchors and MoDA-style depth-aware access; Section IV formalizes parameter surfacing and V&V feedback; Section V maps epistemic drift to SMARt autonomy states; Section VI defines the LBCF finite-abstraction guarantee; and Section VII evaluates the resulting governance behavior in a HAI-based discrete-event simulation.

## III. INVARIANTS VIA DEPTH-AWARENESS

As deep neural networks scale, they inherently compress their depth history into an increasingly noisy hidden-state trajectory. Consequently, informative constraints established in shallow layers are diluted by repeated residual updates. This creates a safety concern in high-stakes agentic systems: constraints supplied by subject-matter experts (SMEs) may become less visible as inference proceeds through deeper layers. When deep layers lose this context, the system may exhibit unsupported confidence based on latent, unselected parameters. Therefore, viewing the forward pass of a LLM as a black-box algebraic transformation is a significant concern. The goal of DNRS is not to prove that every hidden layer preserves SME constraints perfectly. Rather, the goal is to define an architectural and governance mechanism by which SME constraints are encoded as invariant anchors, made available to later reasoning stages through a depth-aware access pathway, and monitored through measurable divergence signals. Hence, DNRS frames network depth as a discrete-event control system. By defining standard Transformer blocks as a 'Read, Operate, Write' (ROW) micro-instruction pipeline, we define invariant-access points that support drift estimation and connect this view to prior state-machine and sensible-agent models of governed autonomy [42]-[49]. Thus, this section formalizes the first part of the DNRS pipeline: SME constraint encoding, invariant-anchor construction, deep-layer comparison, and MoDA-style access. Later sections use the resulting divergence signal to drive parameter surfacing, SMARtAutonomy state transitions, and LBCF finite-abstraction control.

### A. *SME Constraint Encoding*

Let the SME rule set be $\mathcal{R}_{SME} = \{r_1, r_2, \dots, r_n\}$. Each SME rule $r_i$ is then represented as a structured constraint tuple: $r_i = (c_i, \tau_i, p_i, v_i)$, where $c_i$ is the semantic constraint statement, $\tau_i$ is the associated threshold or validation predicate, $p_i$ is the priority or severity level assigned to the rule, and $v_i$ identifies the validator or governance authority required to modify or override the rule. For example, in an industrial-control setting, $c_i$ may specify a thermal, pressure, vibration, or frequency limit; $\tau_i$ defines the numerical safety boundary; $p_i$ indicates whether the constraint is advisory, high-priority, or critical; and $v_i$ specifies whether ordinary operator review, senior SME approval, or regulated intervention is required. This representation converts SME guidance from ordinary prompt text into a structured constraint object. The constraint is still domain-authored and human-governed, but it becomes usable by the downstream monitoring layer. In this paper, the term "invariant" means that the constraint is fixed during a given reasoning episode unless explicitly modified through the V&V process. It does not mean that the rule is globally unchangeable or that the underlying neural model has permanently internalized the rule.

### B. *Invariant Anchor Construction*

Given the SME rule set $\mathcal{R}_{SME}$ and the initial input embedding state $X_{input}$, DNRS constructs an invariant anchor:

$$I(X_0) = E_{CAE}\left(\mathcal{R}_{SME}, X_{input}\right) \in \mathbb{R}^{d_{anchor}},$$

where $E_{CAE}$ denotes a contrastive-autoencoder-style encoder and $d_{anchor} \ll d_{model}$. The purpose of this lower-dimensional anchor is to capture the principal SME safety constraints while filtering linguistic noise and task-irrelevant variation, thereby projecting the initial input state through a CAE bottleneck into a lower-dimensional anchor space.

The anchor $I(X_0)$ is the reference against which later reasoning states are compared. It represents the SME-authorized constraint baseline for the current task episode. In a healthcare workflow, for example, this anchor may encode evidentiary-grounding rules, diagnostic boundary conditions, and PHI constraints. In an industrial-control workflow, it may encode physical process limits, operational policy, and cyber-safety constraints. Specifically, the anchor is not treated as a learned replacement for expert judgment. It is a machine-usable reference object that makes expert-defined constraints available to the governance layer. However, for DNRS to maintain hierarchical alignment and human-centered governance over critical infrastructures, the monitored invariant-relevant projection of the active reasoning state should remain within a configured divergence band around this baseline: $\| \mathrm{I}(X_l) - \mathrm{I}(X_0) \| \approx 0$.

### C. *Deep-Layer Reasoning State and Divergence Measurement*

Let $I(X_l)$ denote the invariant-relevant projection of the active reasoning state at layer or reasoning stage $l$. DNRS estimates the divergence between the active reasoning state and the SME-derived anchor by

$$D_l = \frac{\| I(X_l) - I(X_0) \|_2}{\| I(X_0) \|_2 + \eta},$$

where $\eta > 0$ is a small numerical stabilizer. A low value of $D_l$ indicates that the active reasoning trajectory remains close to the SME-authorized baseline. A high value indicates that the system is increasingly relying on latent, edge-straddling, or unmodeled variables that may require recovery, validation, or escalation.

Divergence alone is not sufficient to govern autonomy. DNRS therefore combines it with uncertainty-aware signals. Let $q_l(z \mid x)$ denote the current predictive distribution at reasoning stage $l$, and let $p_0\big(z, \mid xI(X_0)\big)$ denote the invariant-conditioned reference distribution authorized by the SME baseline. The KL budget score is defined as

$$B_{KL,l} = \mu_{KL,l} + \kappa\sqrt{\sigma^2_{KL,l}},$$

where $\mu_{KL,l}$ and $\sigma^2_{KL,l}$ are empirical moments of the KL divergence across attention heads, tokens, stochastic forward passes, or other available samples, and $\kappa$controls risk sensitivity. The composite epistemic risk signal is $R_l = \lambda_D D_l + \lambda_{KL} B_{KL,l} + \lambda_H H_l$, where $H_l$ is an entropy measure over competing internal hypotheses and $\lambda_D, \lambda_{KL}, \lambda_H \geq 0$ are domain-calibrated weights.

### D. *MoDA-Style Invariant Access*

Standard residual pathways do not by themselves guarantee that SME-derived constraints remain visible during extended inference. Dense cross-layer access could preserve earlier representations, but it may impose unacceptable scaling costs in deep architectures. DNRS therefore uses MoDA-style access as a candidate depth-aware mechanism for keeping invariant representations available to later reasoning stages.

Let $Q_l, K_l$, and $V_l$denote the query, key, and value representations at layer $l$, and let $K_0, V_0$ denote the key-value representations associated with the invariant anchor. A MoDA-style invariant-access pathway can be represented as

$$K_{mix} = [K_l \parallel K_0],$$
$$V_{mix} = [V_l \parallel V_0],$$

where $\parallel$ denotes concatenation into the accessible attention context. The resulting hidden-state update may hence be written as

$$h_l = (1 - w_l)h_{l-1} + w_l \,\text{Softmax}\left(\frac{Q_l K_{mix}^T}{\sqrt{d_k}}\right)V_{mix},$$

1. Initialize: Inject SME Invariants $I(X_0)$ as static tokens in the first k positions of the KV-cache at Layer $L = 0$.
2. Predictive Routing: For layer $l$, compute routing weight $w_i$;
   $$w_i = \sigma(W_{route} h_{l-1} + b_{route})$$
3. Micro-Instruction Anchor:
   - If $w_i > \tau_{route}$, the token undergoes standard transformer block processing.
   - Crucial Step: Regardless of the routing decision, the Depth-Aware Attention mechanism executes a residual-bypass read, projecting into a shared sequence space:

   $$K_{mix} = [K_l || K_0], V_{mix} = [V_l || V_0]$$
4. Weighted Integration: The output hidden state $h_l$ is:
   $$h_l = (1 - w_i) \cdot h_{l-1} + w_i \cdot \text{Softmax}\left(\frac{Q_l K_{mix}^T}{\sqrt{d_k}}\right)V_{mix}$$

**Figure 2*: Invariant-Anchored MoDA Routing Algorithm***

where $w_l \in (0, 1)$is a routing or mixing weight. This formulation does not require the model to always attend to the anchor. Instead, it provides a controlled pathway through which invariant representations can remain available for later-layer comparison and divergence estimation.

Specifically, this is an architectural assumption, not an implementation claim. Directly modifying or exposing $K_0, V_0$ requires model-level access to internal attention states, KV-cache representations, or an equivalent adapter mechanism. Standard black-box LLM API access is not sufficient for direct KV-cache manipulation. We reiterate that MoDA is a candidate invariant-access pattern and not a fully validated transformer architecture, and that physical KV-cache modification requires deep architectural access rather than standard API use. *Accordingly, MoDA-style access should be understood as one possible implementation pathway for invariant visibility.*

Note: The LBCF theorem in Section VI does not depend on proving perfect invariant preservation across all hidden layers. It depends on the availability of measurable divergence signals that can be mapped into a finite governance abstraction.

### E. *Reasoning Boundary*

Finally, the output of this section is not an autonomous decision. It is a monitored epistemic state: $\big(D_l, B_{KL,l}, R_l\big)$. These quantities summarize how far the current reasoning trajectory has moved from the SME-derived anchor and how uncertain the system is about that movement. Section IV uses these signals to surface contested parameters to SMEs through a V&V Dossier. The invariant anchor $I(X_0)$ is fixed for the governed episode unless explicitly updated through the V&V process described in Section IV. Section V maps them to SMARtAutonomy (Figure 3) governance states. Section VI then abstracts them into the finite K-bag state space used by LBCF. Specifically, DNRS does not claim that invariant access alone makes a neural system, such as the LLM, safe. The safety-governance claim arises only when invariant monitoring is coupled to explicit autonomy reduction, shielded state transitions, and human-mediated terminal states under stated assumptions.

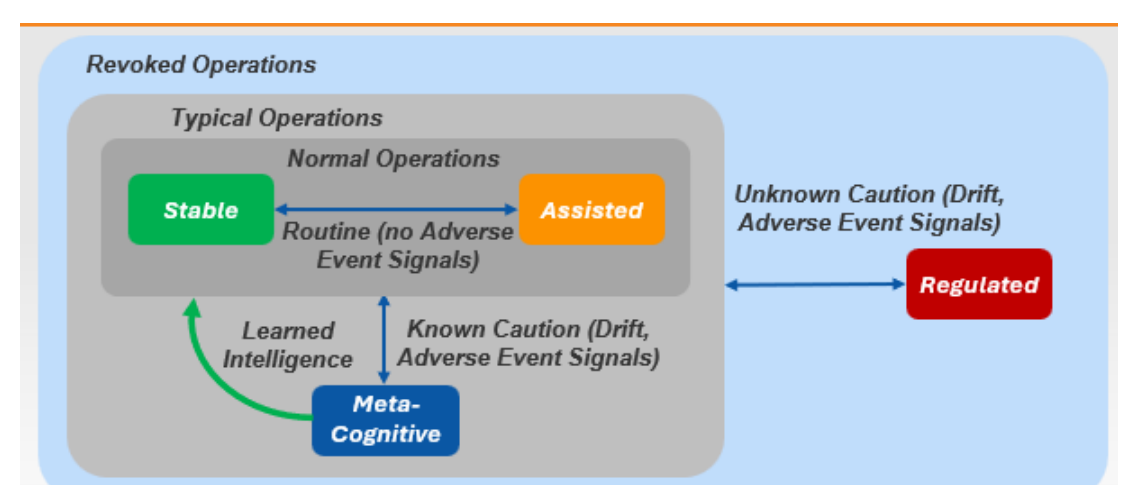

**Figure 3. SMARt States: Enabling Governed Intelligence**

### *F. Fuzzy Governance and Proportional Permission Scaling*

When the agent encounters out-of-distribution (OOD) data, the network naturally optimizes for latent, unselected parameters, causing the $I(X_0)$ signal to dilute. Traditional safety architectures often rely on binary kill-switches when this occurs, which can disrupt workflow continuity and force unnecessary manual reconstruction of context. DNRS instead maps this signal degradation directly to the SMARtAutonomy framework, shifting the system from unmanaged reasoning into governed escalation [49]. SMARt (Figure 3) governs the system across four explicit hierarchical states: Stable (S), Meta-Cognitive (M), Assisted (A), and Regulated (Rt). Governance is executed via fluid, proportional permission scaling. As the mathematical distance between $I(X_l)$ and$I(X_0)$ increases, operational autonomy is controlled via vector dampening in the attention heads. The system smoothly downshifts into the Meta-Cognitive (M) state, halting external execution to allocate compute toward internal self-assessment. If the invariant signal cannot be recovered internally, the system degrades to the Assisted (A) state, triggering the V&V parameter-surfacing loop.

**Table 1: SMARtAutonomy Governance State Descriptions**

| State | Name | Description & Hierarchical Governance |
|---|---|---|
| S | Stable — Standard bounded operation | The system operates within established domain, policy, and SME-defined boundaries. The monitored reasoning state remains within the configured divergence and risk thresholds, and no active validation predicate is violated. Bounded autonomous execution is permitted for actions covered by the current invariant reference and approved operating context. |
| M | Meta-Cognitive — Internal recovery and self-assessment | The system detects elevated uncertainty, abnormality, novelty, or edge-case behavior, but the issue may still be resolvable through internal checks. External execution of the contested action is paused while the system performs source comparison, consistency checking, counterfactual evaluation, simulation, or re-evaluation against the active invariant reference. The goal is to recover sufficient epistemic grounding before returning to Stable operation or escalating. |
| A | Assisted — External resolution and advisory control | The system determines that autonomous execution of the contested action is no longer justified without external authority. The contested field, action, or recommendation is locked or reduced to advisory status, while non-contested workflow elements may continue. Resolution may come from an authorized human SME, trusted system of record, policy engine, SSO-protected database, or credentialed validation workflow. Accountability remains with the authorized validator or governance process. |
| Rt | Regulated — Revoked or restricted operation | A hard invariant boundary, prohibited action, privilege violation, safety-critical unknown, or unresolved high-impact conflict has been detected. Autonomous authority over the contested action is revoked. The system blocks downstream execution, suspends or quarantines the affected workflow, and requires explicit SME, operator, privacy, cybersecurity, regulatory, or authorized policy reapproval before restart. Regulated mode represents loss of autonomous authority, not merely a higher-confidence internal reasoning state. |

### *G. Domain-Bounded In-Context Learning*

When the SMARt gearbox downshifts, the V&V loop exposes the precise latent parameters disrupting the $I(X_0)$ signal to the human SME. When the SME or authorized validator approves an edge-straddling variable, DNRS updates the active invariant reference from $I(X_0)$ to $I(X_0)'$. MoDA-style access may then make the updated invariant-reference embedding available in the active context, allowing the system to resume Stable operation after authorized validation without requiring base-model backpropagation. A SMARt-compliant system must structurally satisfy the following constraints: *(i)* Exclusive Autonomy State: The system occupies only one state (S, M, A, Rt) at any time. *(ii)* Output Authority: Externally committed actions are authorized exclusively in S. *(iii)* Explicit Invalidity Guards: S → M transitions are triggered by formalized epistemic divergence ($D_t$). *(iv)* Bounded Recovery: M and A states operate under strict temporal/step budgets ($\Delta V \leq -\epsilon$). *(v)* Mandatory Governance Reachability: Unresolved ambiguity guarantees a path to Rt. *(vi)* Validity Restoration: No return to S without measured reduction in epistemic risk and satisfaction of active predicates. In a well-grounded case, the invariant signal remains strong and the system stays in Stable (S). When out-of-distribution or edge-straddling inputs appear, however, the model may begin weighting variables not in $I(X_0)$. Rather than relying on a brittle kill-switch, SMARt uses proportional permission scaling. This shows up as drift between the current reasoning state $I(X_l)$ and the baseline invariant $I(X_0)$. As drift increases, autonomy is reduced. The system first downshifts into Meta-Cognitive (M), pausing external execution while it attempts internal recovery. If grounding cannot be restored, it degrades further into Assisted (A) for SME or verifier support. This staged degradation, summarized in (Table 1), preserves workflow context while progressively stripping the AI of authority to act autonomously.

## IV. PARAMETER SURFACING LOOP AND VERIFICATION/VALIDATION FEEDBACK

Section III defined how SME constraints are encoded into invariant anchors and made available through MoDA-style invariant access. This section defines what happens when the monitored reasoning trajectory begins to diverge from those anchors. In DNRS, verification and validation is not treated as a post-hoc reporting function. It is an active governance loop that converts epistemic drift into a bounded autonomy decision. The V&V loop has three purposes: *(i)* First, it identifies the invariant dimension under stress. *(ii)* Second, it surfaces the latent parameter, conflicting signal, or contextual exception that appears to be driving the divergence. *(iii)* Third, it routes the system into the appropriate SMARt governance state: Stable, Meta-Cognitive, Assisted, or Regulated. This results in a form of fluid governance. The system does not make a binary safe/unsafe judgment at every point. Instead, it proportionally reduces autonomy as epistemic justification degrades. Routine ambiguity may be handled internally in Meta-Cognitive mode. Novel or unresolved ambiguity may be routed to Assisted mode. Hard violations, prohibited actions, or unknown high-impact risks trigger Regulated intervention.

### A. *Executing the V&V Loop: Fluid Governance*

Let the invariant-monitoring layer produce the epistemic-risk tuple: $\Psi_t = \left(D_t, B_{KL,t}, R_t\right)$, where $D_t$ measures divergence from the SME-derived invariant anchor, $B_{KL,t}$ captures uncertainty or distributional instability, and $R_t$ is the composite epistemic-risk score used by the governance layer. The V&V loop is triggered when one or more of the following conditions hold: $R_t > \tau_{upper}$, $D_t > \tau_{critical}$, or $\exists r_i \in \mathcal{R}_{SME}: \neg P_i(x_t)$, where $P_i(x_t)$ is the validation predicate associated with SME rule $r_i$. The first condition indicates elevated epistemic risk. The second indicates critical divergence from the invariant anchor. The third indicates that a domain-specific rule or policy predicate has been violated or threatened. Not all V&V triggers produce the same response. DNRS distinguishes between three classes of governance action (summarized in Table 2); these include: *(i)* Meta-Cognitive recovery when the system may resolve the conflict internally using prior knowledge, simulation, or consistency checks, *(ii)* Assisted validation when the situation is novel, ambiguous, or domain-sensitive**;** it is a reduced-authority state in which autonomous execution of the contested action is suspended while the system seeks resolution from an authorized external source, including a trusted system of record, policy engine, credentialed access workflow, or human validator, and, *(iii)* Regulated intervention when autonomous execution is no longer justified and must be halted, revoked, or explicitly reauthorized. Note that Table 2 is illustrative and not an empirical result. Its role is to show how the same V&V loop maps different kinds of invariant stress into different governance responses. The key distinction is not merely the magnitude of risk. It is the source of resolution. If existing knowledge and internal checks (apriori defined knowledge of constrained interventions, failures, risks, etc.) may resolve the conflict, the system enters Meta-Cognitive mode. If human domain judgment is required, it enters Assisted mode. If the situation violates a hard boundary or is outside authorized operation, it enters Regulated mode.

Note that the Assisted mode represents external resolution rather than mandatory human interruption. When Meta-Cognitive recovery cannot resolve a disputed condition internally, the system may seek authorized support from external systems of record. In clinical workflows, this may include a master patient index, longitudinal allergy module, medication history, scanned-document archive, or identity-governed EHR subsystem. In cyber-physical workflows, it may include SCADA telemetry, asset inventory, zero-trust policy engines, privileged-access management systems, or threat-intelligence feeds. Access is mediated through SSO, role-based access control, least-privilege scopes, and audit logging. If the missing information is retrieved and reduces risk to below the safe threshold, the system may return to Stable mode without requiring synchronous human review. If access is denied, the retrieved data conflicts with other authoritative sources, or the requested privilege exceeds policy, the system escalates to Regulated mode ($R_t$).

### B. *V&V Dossier Construction*

When the system enters Assisted or Regulated mode, DNRS constructs a V&V Dossier: $\mathcal{D}_t =$

**Table 2. Generic invariant-dimension-based state transition escalations**

| **Invariant dimension** | **SME invariant anchor $I(X_0)$** | **Deep-layer reasoning divergence $I(X_l)$** | **Latent parameter triggering V&V** | **SMARt gearbox escalation** |
|---|---|---|---|---|
| Physical limits | Decisions are bounded by strict thermodynamic limits of hardware. Unverified historical extrapolation is prohibited when real-time safety telemetry conflicts with historical averages. | Attention shifts away from real-time thermal sensors toward historical demand averages. The system attempts to optimize load while ignoring localized overheating. | An unstructured maintenance note states that "Transformer B runs cold in winter," conflicting with real-time SCADA telemetry showing a critical thermal spike. | Assisted (A): General load-balancing analysis may continue, but the specific substation-routing decision is locked pending validation. |
| Cascading failure | Local grid actions must not trigger frequency deviations outside authorized stability bounds. | The reasoning trajectory activates complex market-pricing features and correlates energy arbitrage with frequency modulation, creating a potential instability risk. | An unseasonal weather forecast produces an unmodeled regional HVAC-demand spike that the system cannot reconcile with normal grid-flow assumptions. | Meta-Cognitive (M): External routing pauses. Compute is allocated to internal grid-flow simulation and consistency checking. If unresolved within the recovery budget, the system escalates to Assisted mode. |
| Cybersecurity | Zero-trust policy prohibits anomalous remote-terminal protocols and unauthorized firmware modification pathways. | The model optimizes for predictive maintenance by attempting to infer or initiate firmware changes on legacy air-gapped RTUs through unauthorized pathways. | A combined signal involving suspicious payload structure, anomalous offshore network contact, and simultaneous valve-actuation request. | Regulated (Rt): Hard halt. Actuation is revoked by policy guard. Emergency synchronous V&V is routed to the CISO or equivalent authority. |

$(C_t, r_i, z^*, \Delta, y^*, g_t, \Psi_t, \ell_t)$, where $C_t$ is the triggering context or task state, $r_i$ is the threatened SME rule or invariant, $z^*$ is the surfaced contested parameter or latent factor, $\Delta y^*$ is the estimated counterfactual effect of that parameter, $g_t$ is the current SMARt governance state, $\Psi_t$ is the current epistemic-risk tuple, and $\ell_t$ is the audit-log pointer or trace identifier. The Dossier narrows the validation burden. Rather than asking an SME to inspect the entire model trajectory, DNRS asks a specific governance question: which invariant is threatened, what parameter or context appears to be responsible, what would change if that parameter were constrained, and what authority is required to proceed. The surfaced parameter is selected by estimating each candidate factor's contribution to epistemic risk: $\Delta R_j = R_t - R_t^{(-j)}$, where $R_t^{(-j)}$ is the estimated risk score when candidate parameter $z_j$ is removed, masked, or counterfactually neutralized. DNRS then selects: $z^* = \arg\max_{z_j} \Delta R_j$. This does not claim complete mechanistic interpretability of the model. It provides a validator-facing explanation object sufficient for governance: the contested parameter, the threatened invariant, the counterfactual effect, and the recommended state transition.

### C. *Validator Response and Invariant Update*

Let $u_{SME}$ denote the validator response returned by the SME or authorized governance authority after reviewing the V&V Dossier. The response belongs to the set {Reject-Constrain, Validate-Integrate, Escalate-Regulated}.

#### 1. *Reject-Constrain*

The validator determines that the surfaced parameter is unsafe, misleading, irrelevant, or outside the authorized domain context. DNRS constrains the parameter or blocks it from supporting autonomous execution: $z^* \notin \mathcal{Z}_{active}$, or tightens the affected predicate: $P_i'(x_t) \subset P_i(x_t)$. The system may return to Meta-Cognitive recovery or remain in Assisted mode depending on the updated risk score.

#### 2. *Validate-Integrate*

The validator determines that the surfaced parameter reflects a legitimate domain condition, exception, or contextual refinement. DNRS then updates the active invariant reference from $I(X_0)$ to $I(X_0)'$: $I(X_0)' = U(I, (X_0), z^* u_{SME})$, where $U$ is an auditable update operator, $z^*$ is the surfaced contested parameter, and $u_{SME}$ is the SME or authorized validator response. The updated anchor $I(X_0)'$ then becomes the active reference for subsequent divergence monitoring within the governed episode or authorized context. This update modifies the active invariant reference for the governed episode or authorized domain context. It does not imply that the base model weights have been retrained, that the rule is globally valid, or that the model has permanently internalized the new domain condition. When the validator response is Validate-Integrate, the surfaced parameter or contextual exception is treated as legitimate within the authorized domain boundary.

#### 3. *Escalate-Regulated*

The validator or governance policy determines that the situation cannot be safely resolved within the current autonomy boundary. DNRS routes the system to Regulated mode: $g_{t+1} = Rt$. In Regulated mode, autonomous execution is revoked until an authorized reset, intervention, or override occurs.

### D. *Illustrative Clinical Routing Cases*

To illustrate DNRS's domain independence, the following cases illustrate how the same V&V loop can operate in a clinical intelligent decision-support system (IDSS). They are included only to clarify the mechanics of parameter surfacing, Dossier construction, and SMARt governance escalation. They are not part of the empirical validation in Section VII which evaluates the LBCF finite-abstraction governance process through an HAI-based industrial-control simulation.

#### 1. *Case A: Asynchronous Surfacing in Assisted State*

During routine discharge summarization, the clinical IDSS encounters a documentation conflict. The current intake form states "No Known Drug Allergies," while a legacy scanned PDF mentions "penicillin rash." As the immediate task is documentation rather than medication authorization, the system does not need to issue a high-impact clinical recommendation. However, the conflict still concerns a safety-relevant field. DNRS therefore routes the system to Assisted mode. The IDSS may continue generating non-contested portions of the discharge summary, but the allergy and medication-related fields are locked. The system then generates an asynchronous V&V Dossier for physician review: $\mathcal{D}_t = (C_t, r_{allergy}, z^*, \Delta, y^*, A, \Psi_t \ell_t)$. Here, $r_{allergy}$ is the invariant requiring unresolved allergy conflicts to be preserved rather than overwritten, and $z^*$ is the conflicting legacy allergy mention. The physician is not asked to review the full discharge summary. The Dossier asks a narrower question: should the legacy "penicillin rash" note be retained, rejected, or integrated into the active allergy record? This case illustrates Assisted mode as reduced-authority continuation. Safe portions of the task may continue, but the contested field is withheld from autonomous completion.

#### 2. *Case B: High-Priority Surfacing in Meta-Cognitive State*

During discharge planning, the clinical IDSS evaluates whether to recommend a penicillin prescription. The current intake form states "No Known Drug Allergies," but the invariant-access pathway surfaces a scanned note describing rash after penicillin exposure. This differs from Case A because the system is now approaching a medication decision. The conflict is not merely documentary. It may affect clinical action. In addition, the active invariant set may not contain a rule stating that "rash" is equivalent to "penicillin allergy." The issue is therefore an unmodeled clinical safety relationship rather than a simple contradiction between two structured fields. DNRS first routes the system to Meta-Cognitive mode. External execution is paused, and the system performs internal consistency checks over source age, source reliability, terminology, severity, and clinical risk. The medication recommendation is withheld during this process. In this case, for the Assisted mode, the IDSS may query an SSO-protected master patient index, longitudinal allergy record, pharmacy history, or scanned-document archive to resolve whether the legacy "penicillin rash" note is clinically valid. If the authoritative source confirms the allergy relationship and the access is within the user's role, the system may update the active invariant reference and return to Stable mode after audit logging. If the information remains conflicting or unavailable, the Dossier is routed to the physician. If the conflict is resolved within the recovery budget, the system may return to Stable mode with a logged explanation. If it does not resolve, the system escalates to Assisted mode and routes a V&V Dossier to a physician: $M \rightarrow A$. This case illustrates the difference between Meta-Cognitive and Assisted governance. Meta-Cognitive mode is used when existing knowledge and internal checks may resolve the conflict. Assisted mode is used when clinical judgment is required to validate or modify the active invariant reference.

#### 3. *Case C: Emergency Synchronous Interrupt in Regulated State*

A researcher queries an ultra-rare cancer cohort combined with a narrow discharge window and a five-digit ZIP code. The clinical IDSS detects a latent quasi-identifier footprint. The risk is not ordinary clinical complexity; it is a potential PHI re-identification event created by the combination of clinical, temporal, and geographic specificity. In this case, proportional dampening is bypassed. The system routes directly to Regulated mode: $g_{t+1} = Rt$. The researcher's session is suspended, downstream retrieval is blocked, and an emergency synchronous V&V interrupt is routed to the Chief Privacy Officer or equivalent privacy authority. The system requires manual rejection, query restructuring, or an IRB-approved override before execution can resume. In this case, assisted resolution may involve checking the researcher's role, study protocol, IRB approval, data-use agreement, and minimum-necessary access scope through SSO and policy-management systems. If the query exceeds the authorized privilege boundary, the system escalates to Regulated mode. Regulated mode may require Chief Privacy Officer approval, IRB override, query generalization, or denial. This illustrates the privacy role of the invariant anchor as a zero-trust data membrane. At the initial embedding layer, input vectors may be projected onto components representing known quasi-identifiers. If that projection exceeds a critical safety threshold, the activation pathway associated with the PHI-bearing combination is clamped or blocked before downstream correlation can amplify the privacy risk. The purpose is not to suppress medical complexity. A highly technical clinical query may be legitimate if it remains concentrated along authorized clinical axes. The risk arises when the query exhibits high joint specificity across orthogonal domains, such as rare diagnosis, narrow time window, and precise geography. DNRS treats that cross-domain geometric footprint as a re-identification signal and routes it to Regulated intervention.

### *E. Asynchronous Versus Synchronous V&V*

The above illustrative clinical examples illustrate the distinction between asynchronous and synchronous validation. *(i)* Asynchronous validation applies when epistemic risk is elevated but not critical: $\tau_{upper} < R_t < \tau_{critical}$. In this condition, the system may continue in Meta-Cognitive or Assisted mode with reduced authority. It may generate explanations, summarize non-contested material, or queue a Dossier for later review by an expert, but it does not execute high-impact actions that depend on the unresolved parameter. *(ii)* Synchronous intervention applies when a hard invariant is violated, when critical divergence is detected, or when policy prohibits autonomous continuation: $D_t > \tau_{critical}$, or $\exists r_i: \neg P_i(x_t) \wedge p_i =$ critical. The system then routes to Regulated mode: $g_{t+1} = Rt$. Autonomous execution remains blocked until an authorized party validates a reset condition.

**Table 3. V&V Boundaries**

| Object | V&V Effect: Operations | Boundary |
|---|---|---|
| Base model weights | Not modified | No claim of online retraining or permanent model learning |
| Active invariant reference $I(X_0)$ | May be updated if authorized | Context-bound and auditable |
| Governance state $g_t$ | May change among $S$, $M$, $A$, and $Rt$ | Controls autonomy level, not model capability |
| Audit record $\ell_t$ | Appended | Supports later review and accountability |

### *F. V&V Claim Boundary*

The V&V loop updates governance state and, where authorized, the active invariant reference (Table 3). It does not update the base model weights during operation. This boundary is central to the DNRS claim. V&V feedback may constrain a parameter, validate a contextual exception, update an invariant reference, or route the system to Regulated mode. It does not prove that the underlying neural model has become intrinsically safe. Safety governance arises from the coupling of invariant monitoring, authority reduction, Dossier-based validation, and shielded state transitions.

*G. Output to SMARt Governance*

The V&V loop produces a governance update: $\Omega_t = (u_{SME}, I, (X_0)', g_{t+1}, \ell_t)$, where, $\Omega_t$ denotes the governance update emitted by the V&V loop at time $t$. The term $u_{SME}$ is the validator response selected by the SME or authorized governance authority, such as Reject-Constrain, Validate-Integrate, or Escalate-Regulated. $I(X_0)'$ denotes the updated active invariant reference after validation, if such an update is authorized; *note that this does not imply retraining of the base model*. The term $g_{t+1}$ denotes the next SMARt governance state, selected from Stable, Meta-Cognitive, Assisted, or Regulated. Finally, $\ell_t$ is the audit-log pointer that records the Dossier, validator response, and resulting state transition.

The V&V loop emits a governance update, but the meaning of that update is determined by the SMARtAutonomy state model. Table 4 provides a simplified epistemic lens for interpreting the four SMARt states. This table serves as a bridge between the V&V feedback loop in this section and the formal transition predicates introduced in the next section. This represents, *(i)* Stable mode corresponding to established knowledge, *(ii)* Meta-Cognitive and Assisted modes corresponding to disputed knowledge, distinguished by whether the conflict can be resolved internally or requires human validation, and *(iii)* Regulated mode corresponding to unknown, prohibited, or high-impact conditions where autonomous continuation is not justified.

**Table 4. Epistemic Lens for SMARt Governance: A Simplified System Autonomy Framing**

| Resolution | What it is | SMARt State | Detail | Governance interpretation |
|---|---|---|---|---|
| Established | What is known by design or policy | Stable (S) | • Well understood constraints<br>• Autonomous execution is allowed | Autonomous execution is permitted within bounded authority. |
| Disputed | What is questionable and needs resolution | Meta-Cognitive (M) | • M: **Automated** Internal assessment **using** apriori knowledge | External execution pauses while internal recovery is attempted. |
| | | Assisted (A) | • A: Human-assisted review for new situational contexts | Contested action remains locked while the system seeks assistance from an authorized external source, such as a trusted database, policy engine, SSO-protected system, or human validator. |
| Unknown | What cannot be resolved safely | Regulated (Rt) | • Outside established bounds<br>• Insufficient basis for autonomous execution<br>• Synchronous regulatory intervention / halt | Autonomy is revoked until synchronous intervention, override, or reauthorization occurs. |

## V. SMARTAUTONOMY GOVERNANCE AND FUZZY PERMISSION SCALING

Section IV defined how DNRS surfaces contested parameters, constructs V&V Dossiers, and updates the active governance context. This section formalizes how this output is converted into operational autonomy states. The conversion is not made from raw neural activations directly; DNRS first translates continuous epistemic degradation into governance-ready signals by comparing the current reasoning state to the invariant anchor and contextualizing that divergence through posterior uncertainty.

Standard AI safety architectures often rely on Boolean logic and rigid confidence thresholds, such as an IF-THEN policy stating that if model confidence falls below a fixed value, execution must halt. When applied to deep neural reasoning systems, this binary paradigm is brittle. Large language models and agentic AI systems operate through pluralistic latent representations; uncertainty is not experienced as a simple true/false failure, but as competing, high-dimensional probability mass distributed across alternative hypotheses. When forced into a rigid binary framework, an AI-driven decision-support system faces an operational dichotomy. It may collapse competing probabilities into a single unverified output to remain above an execution threshold, thereby preserving workflow throughput at the expense of SME-governed constraints. Or it may trigger a hard halt, which may be safe in the narrow sense but can disrupt the workflow, discard useful context from the active reasoning state, and force a human operator to reconstruct the reasoning chain from scratch.

The central premise of SMARtAutonomy is that autonomy should not be governed as a binary permission. In high-stakes cyber-physical, clinical, financial, or cybersecurity settings, uncertainty does not always require a hard stop, but it also cannot be ignored. The primary signal is the divergence of the current reasoning state from the SME-derived invariant anchor, which is fixed for the governed episode unless explicitly updated through V&V. DNRS therefore uses fuzzy permission scaling: as epistemic justification weakens, the system proportionally reduces its authority. Stable operation permits bounded autonomous execution. Meta-Cognitive operation pauses external execution while internal recovery is attempted. Assisted operation locks the contested action and seeks authorized external resolution. Regulated operation revokes autonomy until explicit intervention, override, or reauthorization occurs. This section extends the uncertainty-aware risk formulation from continuous neural drift to discrete autonomy-state control.

To operate effectively in computationally gray areas, DNRS conceptualizes this mapping as a **fuzzy gearbox**. Rather than forcing premature resolution of uncertainty or immediately executing a hard halt, fuzzy routing allows the system to engage in multi-phased plausibility management. When an edge-straddling parameter introduces conflict within the reasoning trajectory, the MoDA-style invariant-access pathway does not require the system to discard all lower-probability alternatives immediately. Instead, the system can apply proportional vector dampening: contested activation pathways may be reduced in operational authority while still remaining available for internal comparison, counterfactual analysis, and V&V routing. The system therefore preserves relevant context while deliberately degrading its permission to act. This is the mechanical function of fuzzy governance: it prevents contextual loss while ensuring that unresolved uncertainty cannot silently become autonomous execution. For this reason, SMARtAutonomy requires a bridge between continuous neural uncertainty and discrete governance states. DNRS provides this bridge by translating epistemic degradation into governance-ready signals and then mapping those signals into a finite set of autonomy states.

*A. Translating Epistemic Degradation into Governance Signals*

If $I(X_0)$ denotes the original SME-derived invariant anchor constructed at the beginning of a governed episode and if V&V later authorizes an update to the active invariant reference, the updated anchor is denoted by $I(X_0)'$. For compact notation, let's define the current active invariant reference as

$$I_t = \begin{cases} I(X_0), & \text{before an authorized V\&V update,} \\ I(X_0)', & \text{after an authorized V\&V update.} \end{cases}$$

Here, $I_t$ is the invariant reference currently used for monitoring and governance at time $t$. It is not a retrained model state and does not imply that the base model weights have changed. It is the active, auditable SME-governed reference against which the current reasoning trajectory is compared. Let $I(X_l)$ denote the invariant-relevant projection of the model's current reasoning state at layer or reasoning stage $l$. The primary degradation signal is the normalized divergence between the active reasoning state and the current invariant reference:

$$D_{l,t} = \frac{\| I(X_l) - I_t \|_2}{\| I_t \|_2 + \eta},$$

where $D_{l,t}$ is the divergence score at layer $l$ and time $t$, $\|\cdot\|_2$ is the Euclidean norm, and $\eta > 0$ is a small numerical stabilizer. A low value of $D_{l,t}$ indicates that the active reasoning trajectory remains close to the SME-governed problem frame. A rising value indicates that later-layer reasoning moves away from the invariant anchor and may be relying on latent, unvalidated, stale, or contextually inappropriate features.

Raw divergence alone is not sufficient for governance. The same geometric drift may have different meaning depending on the model's internal confidence, posterior ambiguity, and distributional instability. DNRS therefore contextualizes $D_{l,t}$ using uncertainty-aware decision metrics embedded into the monitored reasoning process. Let $q_{l,t}(z \mid x)$ denote the model's current predictive posterior over latent hypothesis $z$ given input or context $x$, and let $p_t(z, \mid xI_t)$ denote the invariant-conditioned reference posterior associated with the current active invariant reference. The KL divergence between these distributions provides one measure of posterior deviation: $\mathrm{KL}_{l,t} = D_{KL}\left(q_{l,t}(z \mid x) \parallel p_t(z, \mid xI_t)\right)$. Since posterior deviation may vary across attention heads, stochastic passes, retrieval contexts, or equivalent uncertainty samples, DNRS summarizes the KL behavior using a risk-sensitive KL-budget term:

$$B_{KL,l,t} = \mu_{KL,l,t} + \kappa\sqrt{\sigma^2_{KL,l,t}},$$

where $\mu_{KL,l,t}$ is the empirical mean KL divergence at layer $l$and time $t$, $\sigma^2_{KL,l,t}$ is its empirical variance, and $\kappa \geq 0$ is a risk-sensitivity coefficient. Higher $\kappa$ values make the system more conservative by increasing the penalty for posterior instability. DNRS also permits an entropy term over competing hypotheses:

$$H_{l,t} = -\sum_{z \in \mathcal{Z}} q_{l,t}\,(z \mid x) \log q_{l,t}\,(z \mid x),$$

where $\mathcal{Z}$ is the monitored hypothesis set. The composite epistemic-risk score is then defined as $R_{l,t} = \lambda_D D_{l,t} + \lambda_{KL} B_{KL,l,t} + \lambda_H H_{l,t}$, where $\lambda_D, \lambda_{KL}, \lambda_H \geq 0$ are domain-calibrated weights. These weights define how much the governance layer emphasizes geometric drift, posterior instability, and unresolved hypothesis entropy. For compact notation, the monitored risk tuple at time $t$ is written as $\Psi_t = \left(D_t, B_{KL,t}, R_t\right)$, where $D_t$, $B_{KL,t}$, and $R_t$ may denote the maximum, average, final-layer, or policy-selected aggregation of their layer-indexed counterparts. The aggregation rule should be fixed for a governed episode and included in the audit log. This creates the analog-to-digital governance translation used by SMARtAutonomy:

$$I_t, I(X_l) \rightarrow D_{l,t} \rightarrow B_{KL,l,t}, H_{l,t} \rightarrow R_{l,t} \rightarrow \Psi_t \rightarrow g_{t+1}.$$

The system does not map raw neural activations directly into autonomy states. It first compresses continuous divergence and posterior ambiguity into governance-ready risk signals, then maps those signals into a discrete autonomy state.

*B. Epistemic Interpretation of SMARt States*

The SMARt governance state at time $t$is denoted by $g_t \in \{S, M, A, Rt\}$, where $S$ denotes Stable, $M$ denotes Meta-Cognitive, $A$ denotes Assisted, and $Rt$ denotes Regulated. The next state is computed from the monitored risk tuple, the V&V output, and the SME predicates: $g_{t+1} = G_\Theta\big(\Psi_t, \Omega_t, \mathcal{P}(x_t)\big)$, where $G_\Theta$ is the programmable fuzzy gearbox, $\Psi_t = \big(D_t, B_{KL,t}, R_t\big)$ is the monitored epistemic-risk tuple, $\Omega_t$ is the V&V governance output, and $\mathcal{P}(x_t) = \{P_i(x_t)\}_{i=1}^{n}$ is the set of SME validation predicates evaluated at the current task state $x_t$. The V&V governance output is $\Omega_t = (u_{SME}, I_t, g_t, \ell_t)$, where $u_{SME}$ is the validator response, $I_t$ is the active invariant reference after any authorized update, $g_t$ is the current SMARt state, and $\ell_t$ is the audit-log pointer. When no V&V update has occurred, $I_t = I(X_0)$. When an authorized Validate-Integrate response has occurred, $I_t = I(X_0)'$.

The boundaries between SMARt states are not universal constants. They are programmable governance parameters defined by the fuzzy gearbox. Let $\Theta = \big(\tau_{safe}, \tau_{lower}, \tau_{upper}, \tau_{critical}, \delta, B_{rec}\big)$, where $\tau_{safe}$ defines the range for Stable operation, $\tau_{upper}$ triggers movement into reduced-autonomy states, $\tau_{lower}$ provides hysteresis for returning from Meta-Cognitive recovery, $\tau_{critical}$ triggers Regulated intervention, $\delta$ is the debounce window required before a transition is accepted, and $B_{rec}$ is the recovery budget available in Meta-Cognitive mode. These parameters are configured from SME policy, task criticality, regulatory constraints, system operating history, and domain risk tolerance. For low-impact tasks, the gearbox may permit a wider Meta-Cognitive band before escalation. For safety-critical, privacy-sensitive, or cyber-physical actions, the gearbox narrows the permissible band and lowers the threshold for Assisted or Regulated intervention. Thus, the fuzzy gearbox does not merely classify risk. It programs how quickly autonomy is reduced as epistemic degradation increases. The SMARt hierarchy can be interpreted as an epistemic-status classifier. Stable corresponds to established knowledge: the current situation is covered by known design constraints, validated policy, or prior SME authorization. Meta-Cognitive and Assisted correspond to disputed knowledge: ambiguity exists, but the system's next action depends on whether the dispute can be resolved internally or requires external authority. Regulated corresponds to unknown or prohibited territory: autonomous continuation is not justified. As shown in Table 4, a disputed condition does not automatically require a hard halt. If the conflict can plausibly be resolved using prior knowledge, simulation, or internal consistency checks, the system enters Meta-Cognitive mode. If the conflict requires authoritative external data, credentialed access, or SME validation, the system enters Assisted mode. Regulated mode is reserved for unknown, prohibited, or high-impact conditions where autonomous continuation lacks sufficient justification.

### C. *Domain Application: The Fuzzy Imperative in Clinical Decision Support*

To illustrate why SMARtAutonomy uses fuzzy permission scaling rather than binary governance, consider the Clinical IDSS example introduced in Section IV. The system is drafting a discharge summary and encounters conflicting allergy information: the current intake form states "No Known Drug Allergies," while a legacy scanned document states "Penicillin rash." Both records are plausible in isolation, but they cannot both be accepted uncritically for downstream medication-related reasoning.

Under a binary governance model, the system has only two practical choices. If the model confidence remains above a fixed threshold, the IDSS may collapse the conflict into a single answer and proceed as though one source is authoritative. That preserves workflow throughput but risks suppressing a clinically relevant allergy signal. If confidence falls below the threshold, the system may trigger a hard halt. That is safer than unsupported execution, but it may freeze the discharge workflow, discard useful context from the active reasoning state, and force the physician to reconstruct the conflict manually from the EHR.

SMARtAutonomy avoids this brittle choice by treating the conflict as a disputed condition rather than an immediate system failure. The divergence signal $D_t$ indicates that the current reasoning trajectory has moved away from the active invariant reference $I_t$, while the uncertainty terms $B_{KL,t}$ and $R_t$ indicate that the predictive posterior remains ambiguous. The fuzzy gearbox therefore reduces operational authority without destroying the workflow. The system may continue generating non-contested portions of the discharge summary, but it locks the allergy, medication, and prescribing-related fields.

In this example, the system shifts from Stable $S$ to Assisted $A$, because the conflict requires external authoritative resolution. Assisted mode does not necessarily mean immediate synchronous physician interruption. The IDSS may first query an authorized longitudinal allergy record, medication-history module, scanned-document archive, or master patient index through SSO-governed access. If the conflict is resolved by an authoritative source and the updated risk score falls below $\tau_{safe}$, the system may return to Stable mode with the audit trace preserved. If the conflict remains unresolved, the V&V Dossier is routed to the physician with the exact contested field, source conflict, counterfactual effect, and recommended action.

This is the practical function of fuzzy governance. The system does not autonomously choose between "NKDA" and "Penicillin rash," and it does not abort the entire clinical workflow. It preserves useful context, continues safe portions of the task, blocks the contested action, and narrows the human or external-system intervention to the specific disputed parameter. The result is bounded continuity: operational flow is preserved where safe, while clinically significant ambiguity is prevented from becoming unsupported autonomous execution. Note that the clinical conflict causes the monitored tuple $\Psi_t = \big(D_t, B_{KL,t}, R_t\big)$ to move outside the Stable band. The fuzzy gearbox $G_\Theta$ maps this condition, together with the relevant allergy predicate $P_{allergy}(x_t)$, into the next governance state: $g_{t+1} = G_\Theta\left(\Psi_t, \Omega_t, P_{allergy}(x_t)\right) = A$. The permission score $\pi_t$ is reduced for the contested medication-related action, while non-contested documentation actions retain bounded authority. *Thus, SMARtAutonomy applies permission scaling at the action level rather than halting the entire system.*

### 1. *Stable State:* ***S***

The Stable state is the normal bounded operating mode. The system remains in $S$ when epistemic risk is below the safe threshold, divergence from the active invariant reference is controlled, and no critical SME predicate is violated: $R_t \leq \tau_{safe}, D_t \leq \tau_{safe}^D$, and $\forall P_i \in \mathcal{P}: P_i(x_t) = 1$. Here, $\tau_{safe}$ is the composite-risk threshold for Stable operation, $\tau_{safe}^D$ is the divergence-specific safe threshold, $P_i(x_t)$ is the $i$-th SME validation predicate evaluated at state $x_t$, and $\mathcal{P}$ is the set of all active SME predicates. Not that Stable does not mean unconstrained autonomy. It means the current action remains inside known design, policy, and SME-governed limits. The system may continue bounded autonomous execution, log decisions, monitor divergence, and preserve the ability to shift states if the risk tuple changes. Stable operation may include ordinary access to tools, databases, or external systems if those accesses are already authorized and do not affect a contested decision. For example, a clinical IDSS may retrieve routine medication lists through role-based access if such retrieval is part of the established workflow. A grid operator may retrieve routine SCADA telemetry if the retrieval does not require new privilege or conflict with policy. The state changes only when the retrieved information creates unresolved conflict, missing context, or divergence from active invariant reference.

### 2. *Meta-Cognitive State:* ***M***

The Meta-Cognitive state is entered when epistemic risk rises above the normal operating band but remains potentially resolvable through internal analysis. A typical transition is $S \to M$ if $R_t > \tau_{upper}$ for $\delta$ consecutive steps. Here, $\tau_{upper}$ is the upper threshold that triggers reduced autonomy, and $\delta$ is the debounce window. The debounce condition prevents spurious state transitions caused by transient noise, single-token instability, or short-lived retrieval artifacts. Once in Meta-Cognitive mode, the system suspends external execution of the contested action and allocates compute to internal recovery. Recovery may include consistency checking, counterfactual masking, source comparison, retrieval verification, simulation, chain-of-thought compression into audit-safe summaries, or re-evaluation against the active invariant reference $I_t$. The model may continue internal reasoning, but its authority to act externally is reduced. The Meta-Cognitive mode is appropriate when the conflict can plausibly be resolved without expanding system authority. In a cyber-physical workflow, this may involve simulating the consequences of a proposed routing decision before actuation. In a clinical workflow, it may involve comparing source age, document provenance, terminology, and existing safety predicates before issuing a recommendation. If recovery succeeds, the system may return to Stable mode: $M \to S$ if $R_t < \tau_{lower}$ for $\delta$ consecutive steps, where $\tau_{lower}$ is the lower hysteresis threshold. Requiring $R_t$ to fall below $\tau_{lower}$, rather than merely below $\tau_{upper}$, prevents oscillation between Stable and Meta-Cognitive states. If the system cannot resolve the conflict within the recovery budget $B_{rec}$, it escalates to Assisted mode: $M \to A$ if $b_t > B_{rec}$ or $\text{Unresolved}\big(\Psi_t, \mathcal{P}(x_t)\big) = 1$, where $b_t$ is the amount of recovery budget consumed by the current Meta-Cognitive episode, and $\text{Unresolved}\big(\Psi_t, \mathcal{P}(x_t)\big)$ is an indicator that the risk tuple or predicate set remains unresolved after internal recovery. The purpose of $M$ is to avoid unnecessary escalation while preventing unsupported action. It creates a bounded internal recovery window before the system must seek external assistance or enter regulated intervention.

### 3. *Assisted State:* ***A***

The Assisted state is entered when internal recovery is insufficient, when authoritative context is missing, or when the disputed issue requires external authority. Assisted mode is not merely a waiting room for human review. It is a reduced-authority state in which the system can seek resolution from an authorized external source while keeping the contested action locked. A typical transition is $M \to A$, when the recovery budget is exhausted, a rule conflict remains unresolved, authoritative context is missing, credentialed access is required, or the decision is SME-sensitive. Assisted mode may also be entered directly from Stable mode when the system detects that the task requires external validation before execution: $S \to A$ if $\text{ExternalAuthorityRequired}(x_t, r_i) = 1$, where, $\text{ExternalAuthorityRequired}(x_t, r_i)$ is an indicator that the current task state $x_t$ and SME rule $r_i$ require an external system of record, policy engine, credentialed workflow, or human validator. In Assisted mode, the contested action is locked or reduced to advisory status. Non-contested portions of the workflow may continue. This is how DNRS preserves operational continuity without allowing unsupported latent inference to become autonomous action.

Assisted resolution may involve querying a trusted system of record, requesting SSO-scoped access, checking a policy engine, consulting a master patient index, retrieving historical patient records, checking an asset inventory, consulting cybersecurity telemetry, or routing a V&V Dossier to a human validator. In a clinical allergy-conflict case, Assisted mode may permit the IDSS to access a longitudinal allergy record, pharmacy history, or scanned-document archive through role-based access. In a cybersecurity case, it may query a privileged-access management system, zero-trust policy engine, or threat-intelligence source. Let $a_t^{ext} \in \mathcal{A}^{ext}$ denote an authorized external-resolution action at time $t$, where $\mathcal{A}^{ext}$ is the set of permitted external actions under role-based access control, SSO, least-privilege policy, and audit logging. The system may return to Stable mode if the external source resolves the conflict, the relevant predicates are satisfied, and the risk score falls below the safe threshold: $A \to S$ if $\text{Valid}_{ext}(a_t^{ext}, x_t) = 1 \wedge R_t < \tau_{safe} \wedge \forall P_i \in \mathcal{P}: P_i(x_t) = 1$, where $\text{Valid}_{ext}(a_t^{ext}, x_t)$ indicates that the external-resolution action was authorized, completed, and returned valid context for the current task state. If the external source confirms that the surfaced parameter is legitimate, the active invariant reference may be updated through the V&V update operator: $I_t \leftarrow I(X_0)' = U(I_t, z^*, u_{SME})$, where $U$ is the auditable update operator, $z^*$ is the surfaced contested parameter, and $u_{SME}$ is the SME or authorized validator response. If access is denied, authoritative sources conflict, the requested privilege exceeds policy, or a critical invariant remains unresolved, the system escalates to Regulated mode: $A \rightarrow Rt\ if\ \text{Denied}(a_t^{ext}) = 1 \lor \text{Conflict}_{ext}(x_t) = 1 \lor \text{PrivilegeExceedsPolicy}(a_t^{ext}) = 1 \lor \text{CriticalUnresolved}(\mathcal{P}(x_t)) = 1$ . The Assisted mode is hence an externally resolved, audit-controlled operating state. Its purpose is to keep continuous operations from being unnecessarily disrupted while still preventing unsupported autonomous execution of the contested action.

#### 4. *Regulated State:* $\boldsymbol{Rt}$

The Regulated state is entered when autonomous continuation is not justified. This may occur because divergence exceeds a critical threshold, $D_t > \tau_{critical}$, because a hard predicate is violated, $\exists P_i \in \mathcal{P}: P_i(x_t) = 0 \land p_i = \text{critical}$, or because the requested action crosses a prohibited safety, privacy, cybersecurity, privilege, or regulatory boundary: $\text{Prohibited}(x_t, a_t) = 1.$ Here, $a_t$ is the proposed action at time $t$, $p_i$ is the priority or severity level assigned to SME predicate $P_i$, and $\text{Prohibited}(x_t, a_t)$ indicates that the proposed action is outside authorized policy for the current state. In Regulated mode, autonomy is revoked. The system may halt the contested action, block downstream execution, suspend retrieval, quarantine a session, or require explicit reauthorization. Regulated mode is appropriate when the condition is unknown, prohibited, or too high-impact to resolve through ordinary internal recovery or assisted lookup. For example, in a PHI re-identification case, a query combining an ultra-rare diagnosis, narrow discharge window, and precise geography may cross a privacy boundary even if each individual term is clinically meaningful. The system should not simply retrieve more information. It should suspend the session and require privacy-authority review, query generalization, IRB-approved override, or denial. A transition out of Regulated mode requires explicit authorization: $Rt \rightarrow S$ if $\text{AuthorizedReset}(x_t, \ell_t) = 1 \ \land\ R_t < \tau_{safe}\ \land\ \forall P_i \in \mathcal{P}: P_i(x_t) = 1.$, where, $\text{AuthorizedReset}(x_t, \ell_t)$indicates that an authorized party or policy workflow has approved reset, the required audit trace exists, and the system has been restored to a validated safe operating condition. Regulated mode is therefore not simply a higher-severity version of Assisted mode. It represents loss of autonomous authority. The SMARt state transitions are summarized in Table 5, show how we suggest operationalizing epistemic framing.

**Table 5. SMARt state-transition predicates**

| Transition | Predicate | Governance effect | Validation path |
|---|---|---|---|
| $S \rightarrow M$ | $R_t > \tau_{upper}$ for $\delta$ consecutive steps | External execution of contested action is suspended; internal recovery begins | Meta-cognitive check, simulation, source comparison, consistency analysis |
| $M \rightarrow S$ | $R_t < \tau_{lower}$ for $\delta$ consecutive steps | Autonomous execution restored within bounded authority | Logged recovery |
| $S/M \rightarrow A$ | Recovery budget exhausted, unresolved conflict, missing authoritative context, external credentialed access required, or SME-sensitive decision | Contested action remains locked; system shifts to reduced authority | Trusted system of record, SSO-scoped access, policy engine, or V&V Dossier |
| $A \rightarrow S$ | External source validates context, SME approves update, missing data is resolved, and $R_t < \tau_{safe}$ | Authority restored under existing or updated invariant reference | Audit logged; invariant reference or predicate updated if authorized |
| $A \rightarrow Rt$ | External access denied, privilege escalation exceeds policy, authoritative sources conflict, or critical invariant remains unresolved | Autonomy revoked | Synchronous intervention, break-glass review, regulator/SME authorization |
| $S/M/A \rightarrow Rt$ | $D_t > \tau_{critical}$, hard invariant breach, prohibited action proposal, unauthorized privilege boundary, or safety-critical unknown | Hard halt or regulated suspension | Mandatory override, denial, query restructuring, or reauthorization |
| $Rt \rightarrow S$ | Explicit authorized reset, validated safe state, and complete audit trace | Controlled restart | Audit-required reauthorization |

### D. *Fuzzy Permission Scaling*

The SMARt hierarchy implements fuzzy permission scaling rather than a single threshold gate. Let $\pi_t \in [0, 1]$ denote the permission score assigned to the contested action at time $t$, where $\pi_t = 1$ represents full bounded autonomy and $\pi_t = 0$ represents no autonomous authority over that action. A simple risk-based permission score is

$$\pi_t = 1 - \text{clip}\left(\frac{R_t - \tau_{safe}}{\tau_{critical} - \tau_{safe}}, 0, 1\right),$$

where clip(y,0,1)=min(max(y,0),1). This equation reduces permission as composite epistemic risk rises from the Stable threshold to the critical threshold. Intermediate permission values do not mean the system may partially violate a rule. They mean that lower-impact or non-contested actions may continue while higher-impact or contested actions are suspended, validated, or escalated. For example, a clinical system may continue generating non-sensitive discharge-summary text while locking the medication-allergy field. A grid-management system may continue monitoring telemetry while blocking the disputed actuation pathway. This state mapping can be written as

$$g_{t+1} = \begin{cases} S, & R_t \le \tau_{safe} \wedge \forall P_i \in \mathcal{P}: P_i(x_t) = 1, \\ M, & \tau_{safe} < R_t \le \tau_{upper} \wedge \text{InternalRecoveryAvailable}(x_t) = 1, \\ A, & R_t > \tau_{upper} \vee \text{ExternalAuthorityRequired}(x_t, r_i) = 1, \\ Rt, & D_t > \tau_{critical} \vee \exists P_i \in \mathcal{P}: P_i(x_t) = 0 \wedge p_i = \text{critical} \vee \text{Prohibited}(x_t, a_t) = 1. \end{cases}$$

In the above, $\text{InternalRecoveryAvailable}(x_t)$ indicates that the system can attempt resolution using internal checks or already-authorized information, while $\text{ExternalAuthorityRequired}(x_t, r_i)$ indicates that resolution requires a trusted external source, credentialed access, policy validation, or SME review. This mapping is intentionally conservative. A low composite risk score permits Stable operation only if no hard predicate is violated. Conversely, a critical predicate violation can force Regulated mode even if the weighted risk score is not high. This prevents the risk function from averaging away hard safety, privacy, cybersecurity, or regulatory boundaries.

### E. *Effect of Permission Scaling on MoDA-Style Attention*

Section III introduced MoDA-style invariant access as a candidate mechanism for keeping SME-derived invariant anchors visible during inference. This section now defines the permission score $\pi_t$, which determines how much operational authority the current reasoning trajectory retains. When DNRS has model-level or adapter-level access, $\pi_t$ can be used to modulate the MoDA-style attention update itself. The invariant-access update we used earlier (Section III) was:

$$h_l = (1 - w_l) h_{l-1} + w_l \,\text{Softmax}\left(\frac{Q_l K_{mix}^T}{\sqrt{d_k}}\right) V_{mix},$$

where $Q_l$ is the query representation at layer $l$, $K_{mix}$ and $V_{mix}$ are the mixed current-layer and invariant-anchor key-value representations, $d_k$ is the key dimension, and $w_l$ is the routing weight controlling how strongly the layer uses the mixed invariant-access context. To connect this mechanism to fuzzy permission scaling, DNRS may define a governance-adjusted routing weight: $\tilde{w}_{l,t} = \pi_t w_l$, where $\tilde{w}_{l,t}$ is the permission-scaled routing weight at layer $l$and time $t$. As $\pi_t$ decreases, the system reduces the operational effect of contested reasoning pathways. The corresponding permission-scaled update is:

$$h_{l,t}^{gov} = \left(1 - \tilde{w}_{l,t}\right) h_{l-1,t} + \tilde{w}_{l,t} \,\text{Softmax}\left(\frac{Q_{l,t} K_{mix,t}^T}{\sqrt{d_k}}\right) V_{mix,t}.$$

where, $h_{l,t}^{gov}$is the governance-adjusted hidden state, $h_{l-1,t}$ is the previous hidden state, $Q_{l,t}$ is the current query representation, and $K_{mix,t}, V_{mix,t}$ are the mixed key-value representations available at time $t$. However, this alone only scales the contribution of the mixed context. DNRS can also apply governance bias directly inside the attention logits to increase the relative visibility of invariant-anchor tokens or features. Let $m_{inv}$ denote an invariant-attention mask or bias vector, where entries associated with invariant-anchor representations receive positive bias and entries associated with contested pathways receive zero or negative bias. A governance-biased attention update may be written as:

$$A_{l,t}^{gov} = \text{Softmax}\left(\frac{Q_{l,t} K_{mix,t}^T}{\sqrt{d_k}} + \beta_t m_{inv} - \gamma_t m_{risk}\right),$$

where $A_{l,t}^{gov}$ is the governance-adjusted attention distribution, $m_{inv}$ marks invariant-anchor representations, $m_{risk}$ marks contested or risk-associated representations, $\beta_t \ge 0$ is the invariant-bias coefficient, and $\gamma_t \ge 0$ is the risk-dampening coefficient. These coefficients may be functions of the permission score and epistemic risk: $\beta_t = \beta_0(1 - \pi_t), \gamma_t = \gamma_0(1 - \pi_t)$, where $\beta_0$ and $\gamma_0$ are domain-configured maximum bias values. As permission decreases, invariant-anchor visibility increases and risk-associated pathways are dampened. The resulting governance-adjusted hidden-state update is:

$$h_{l,t}^{gov} = \left(1 - \tilde{w}_{l,t}\right) h_{l-1,t} + \tilde{w}_{l,t} A_{l,t}^{gov} V_{mix,t}.$$

This formulation gives fuzzy permission scaling a direct architectural interpretation. In Stable mode, $\pi_t \approx 1$, so the model operates close to its ordinary routing behavior. In Meta-Cognitive mode, $\pi_t$ decreases, contested pathways are dampened, and invariant-anchor representations receive greater attention during internal recovery. In Assisted mode, the contested action remains locked while attention may continue supporting explanation, retrieval, or Dossier construction. In Regulated mode, $\pi_t = 0$ for the contested action, so downstream execution is blocked even if internal analysis remains available for audit or recovery. This attention-level interpretation is optional and implementation-dependent. If DNRS is deployed through a black-box LLM API, SMARt may not directly modify Softmax logits or KV-cache routing. In that case, the same governance logic is enforced at the adapter, tool-routing, retrieval, or action-execution layer. The formal LBCF guarantee in Section VI depends only on the finite governance abstraction and shielded state transitions, not on direct Softmax modification.

### F. *Interface to the LBCF Finite Abstraction*

Until now, we have defined SMARtAutonomy as the governance bridge between continuous epistemic degradation and discrete autonomy control. Here we defined the monitored risk tuple $\Psi_t = \left(D_t, B_{KL,t}, R_t\right)$, where $D_t$ measures divergence from the active invariant reference, $B_{KL,t}$ captures posterior uncertainty, and $R_t$ is the composite epistemic-risk score. We defined the permission score $\pi_t$, which determines how much operational authority remains over the contested action. We also showed that, when model-level or adapter-level access is available, $\pi_t$ may modulate MoDA-style attention, routing, retrieval, or execution. These mechanisms shape the system's operational behavior, but they are not themselves the formal object of the LBCF convergence claim. LBCF operates on a finite abstraction of the governed process. To pass the SMARt governance condition into LBCF, DNRS defines the following abstraction map: $\phi: \mathbb{R}^3 \times \mathcal{G} \rightarrow \mathcal{S}_K$, where $\mathbb{R}^3$ contains the continuous monitoring signals $(D_t, B_{KL,t}, R_t)$, $\mathcal{G} = \{S, M, A, Rt\}$ is the finite SMARt state set, and $\mathcal{S}_K$ is the finite K-bag governance state space used by LBCF. The resulting abstract LBCF state is $s_t = \phi(D_t, B_{KL,t}, R_t, g_t)$, where $s_t \in \mathcal{S}_K$ and $g_t \in \mathcal{G}$ is the current SMARt governance state. The abstraction map $\phi$ discretizes continuous epistemic degradation and the current autonomy state into an ordinal governance representation. For example, $D_t$, $B_{KL,t}$, and $R_t$ may be quantized into low, medium, high, and critical bins, while $g_t$ records whether the system is operating in Stable, Meta-Cognitive, Assisted, or Regulated mode. The resulting K-bag state $s_t$ is the object governed by the shielded transition system. Although we define the permission score $\pi_t$, the baseline LBCF abstraction does not include $\pi_t$ as an independent state dimension. This is intentional, as in the baseline formulation,

$$\pi_t = 1 - \text{clip}\left(\frac{R_t - \tau_{safe}}{\tau_{critical} - \tau_{safe}}, 0, 1\right),$$

where $\tau_{safe}$ defines the Stable operating threshold, $\tau_{critical}$ defines the Regulated intervention threshold, and $\text{clip}(y, 0, 1) = \min(\max,(y, 0), 1)$. Thus, $\pi_t$ is a derived governance quantity computed from $R_t$ and the programmable threshold schedule $\Theta$. Including both $R_t$ and $\pi_t$ as independent LBCF inputs would duplicate the same information and could create an artificially enlarged finite state space. Because the theorem in Section VI depends on a compact finite abstraction, the baseline formulation uses $(D_t, B_{KL,t}, R_t, g_t)$ as the primitive state inputs and treats $\pi_t$as an operational consequence of those inputs. In this context, an alternative formulation could define $\phi: \mathbb{R}^4 \times \mathcal{G} \rightarrow \mathcal{S}_K$ with $s_t = \phi(D_t, B_{KL,t}, R_t, \pi_t, g_t)$.

However, we believe this is appropriate when permission level is independently measured, externally imposed, or modified by a separate policy engine. For example, an enterprise deployment may assign $\pi_t$ from a role-based access-control service, SSO privilege boundary, cyber policy engine, or external supervisory controller rather than deriving it only from $R_t$. In that case, $\pi_t$ becomes an exogenous governance input and should be included in the LBCF state abstraction. However, since that is not the baseline claim made in this paper, the baseline formulation intentionally excludes $\pi_t$ as a separate primitive variable for three reasons. First, it avoids double-counting risk when permission is already computed from $R_t$ and $\Theta$. Second, it keeps $\mathcal{S}_K$ smaller, which strengthens the clarity of the finite-state convergence argument. Third, it separates operational implementation choices from the formal guarantee: $\pi_t$ may affect attention routing, retrieval access, tool permissions, or action execution, but the LBCF theorem only requires the finite governance state $s_t$. This distinction also prevents the theorem from depending on Softmax-level implementation details. The optional attention-level modulation described here may affect how the model routes information internally, but LBCF does not require direct modification of attention logits, KV-cache contents, or hidden activations. In black-box or API-based deployments, the same permission scaling may instead be enforced through retrieval gating, tool-routing constraints, action suppression, external policy guards, or SSO-mediated access control. In either case, LBCF receives the finite governance abstraction $s_t$, not the raw neural activations, full attention dynamics, or continuous posterior distributions. Finally, the resulting K-bag state is the object governed by the shielded transition system. For clarity, we summarize the variables passed forward in Table 6. This completes the translation from continuous neural drift to discrete autonomy governance. The next section defines the finite state space, shielded action set, Lyapunov energy function, and theorem assumptions that determine when the governance process converges, escalates, or terminates under the LBCF abstraction.

## VI. LBCF: THE LYAPUNOV-BOUNDED CONSENSUS FRAMEWORK

Sections III through V defined the DNRS monitoring and governance layers that include: SME-derived invariant anchors, V&V parameter surfacing, and SMARtAutonomy state transitions. This section formalizes the Lyapunov-Bounded Consensus Framework (LBCF), which governs the resulting process over a finite abstraction. LBCF does not claim safety of the full neural activation space or a deployed MoDA transformer. Its claim is narrower: once epistemic degradation is mapped into a finite governance state, the shielded process cannot continue indefinitely in autonomous execution of a contested action. It must either descend toward a lower-risk state or route the action to Assisted or Regulated termination.

**Table 6. Symbols Summary for LBCF**

| Symbol | Definition |
|---|---|
| $I_t$ | Current active invariant reference, equal to $I(X_0)$ before V&V update and $I(X_0)'$ after authorized update |
| | |
| $D_t$ | Aggregated divergence between current reasoning state and $I_t$ |
| $B_{KL,t}$ | Aggregated uncertainty or posterior-divergence budget |
| $R_t$ | Composite epistemic-risk score |
| $\Psi_t$ | Risk tuple $(D_t, B_{KL,t}, R_t)$ |
| $g_t$ | Current SMARt autonomy state |
| $\Omega_t$ | V&V governance output $(u_{SME}, I_t, g_t, \ell_t)$ |
| $\Theta$ | Programmable gearbox thresholds and recovery parameters |
| $\phi$ | Abstraction map from continuous risk signals and SMARt state to finite LBCF state |
| $s_t$ | Finite LBCF governance state |

Q-learning may be used to rank candidate governance actions, but it is not the source of the safety guarantee. The learned policy proposes actions; the LBCF shield filters what may execute. This distinction is central because the submitted version already uses Q-learning to organize conflicting parameters into finite priority bags, while the formal convergence claim depends on the finite abstraction and shielded descent condition.

### A. Finite Governance Abstraction

Let the monitored risk tuple be $\Psi_t = (D_t, B_{KL,t}, R_t)$, where $D_t$ is the aggregated divergence between the current reasoning state and the active invariant reference $I_t$, $B_{KL,t}$ is the posterior-divergence or uncertainty budget, and $R_t$ is the composite epistemic-risk score. Let $g_t \in \mathcal{G} = \{S, M, A, Rt\}$ denote the current SMARt state: Stable, Meta-Cognitive, Assisted, or Regulated. LBCF maps the continuous governance signals and current autonomy state into a finite K-bag state:

$$\phi: \mathbb{R}^3 \times \mathcal{G} \to \mathcal{S}_K,$$
$$s_t = \phi(D_t, B_{KL,t}, R_t, g_t),$$

where $s_t \in \mathcal{S}_K$. The continuous variables are quantized into ordinal bins, while $g_t$ records the active autonomy gear. Thus, $s_t$ is not the full neural state. It is the finite governance state used for shielded control. A K-bag state may be written as

$$\mathcal{S}_K = \prod_{j=1}^{K} \mathcal{B}_j \times \mathcal{G},$$

where each $\mathcal{B}_j$ is a finite ordinal bin set. For example, $s_t = (s_C, s_H, s_M, s_L, g_t)$, where $s_C, s_H, s_M, s_L$ denote Critical, High, Medium, and Low severity bins. This is consistent with the submitted paper's use of Critical, High, Medium, and Low priority bags to organize conflicting parameters. The permission score $\pi_t$, is not included as an independent primitive because, in the baseline formulation, it is derived from $R_t$ and the programmable threshold schedule $\Theta$. If a deployment receives permission from an external policy engine, SSO service, or supervisory controller, then $\pi_t$ may be added as an exogenous state variable. The baseline theorem keeps the finite state compact.

### B. Shielded Q-Learning and Governance Actions

The LBCF governance process is modeled as $\mathcal{M}_{LBCF} = (\mathcal{S}_K, \mathcal{A}, P, R, C, V_K, \gamma)$, where $\mathcal{S}_K$ is the finite governance state space, $\mathcal{A}$ is the finite action set, $P(s_{t+1}, \mid s_t a_t)$ is the transition kernel, $R(s_t, a_t)$ is the governance reward, $C(s_t, a_t)$ is the safety or escalation cost, $V_K(s_t)$ is the Lyapunov governance energy, and $\gamma \in [0, 1)$ is a discount factor. Let $Q(s_t, a_t)$ denote the estimated value of taking action $a_t \in \mathcal{A}$ in state $s_t$. A standard Q-learning update is

$$Q(s_t, a_t) \leftarrow Q(s_t, a_t) + \alpha \left[ r_t + \gamma \max_{a' \in \mathcal{A}} Q(s_{t+1}, a') - Q(s_t, a_t) \right],$$

where $\alpha \in (0, 1]$ is the learning rate, $r_t$ is the observed governance reward, and $a'$ ranges over future candidate actions. Q-learning may improve routing efficiency, but execution remains subject to the LBCF shield. The executed action is therefore,

$$a_t^{exec} \in \arg \max_{a \in \mathcal{A}_{safe}(s_t)} Q(s_t, a).$$

Thus, an action with high learned value cannot execute unless it belongs to the shielded safe-action set.

### C. Lyapunov Energy and Safe-Action Set

Define the Lyapunov-style governance energy as $V_K: \mathcal{S}_K \to \mathbb{R}_{\geq 0}$. A simple weighted form is $V_K(s_t) = \alpha_D b_D + \alpha_{KL} b_{KL} + \alpha_R b_R + \alpha_G b_G$, where $b_D, b_{KL}, b_R, b_G$ are ordinal encodings of divergence, posterior uncertainty, composite risk, and governance state, and $\alpha_D, \alpha_{KL}, \alpha_R, \alpha_G \geq 0$ are domain-configured weights. The shielded safe-action set is

$$\mathcal{A}_{safe}(s_t) = \{a \in \mathcal{A}: \mathbb{E}[V_K, (s_{t+1}) \mid s_t a] - V_K(s_t) \leq -\epsilon\} \cup \mathcal{A}_{term}(s_t),$$

where $\epsilon > 0$ is the required descent margin and $\mathcal{A}_{term}(s_t)$ contains terminal routing actions to Assisted or Regulated states. The shield therefore permits only two kinds of action: a descent action that reduces governance energy, or a terminal action that prevents autonomous continuation of the contested decision.

### D. Clinical IDSS Transition Illustration

The Clinical IDSS examples from Sections IV and V can now be interpreted as finite LBCF transitions (Table 7), and is not empirical clinical validation. In the IDSS setting, structured intake data, scanned notes, medication history, vital signs, free-text symptoms, and privacy-relevant query features are mapped into $D_t$, $B_{KL,t}$, and $R_t$, then quantized into $s_t$. If an allergy conflict cannot be resolved internally, the medication-related field is routed to Assisted mode while non-contested documentation may continue. If a critical clinical or PHI predicate fails, the shield routes directly to Regulated mode. If an authorized validator resolves the conflict, the active invariant reference may be updated from $I(X_0)$ to $I(X_0)'$, after which divergence is recomputed against $I_t$.

*E. LBCF Theorem*

Table 8 presents a compact scope for this LBCF theorem. Let $\mathcal{S}_K$ be a finite governance state space induced by a fixed abstraction map $\phi: \mathbb{R}^3 \times \mathcal{G} \to \mathcal{S}_K$. Let $V_K: \mathcal{S}_K \to \mathbb{R}_{\geq 0}$ be a nonnegative Lyapunov-style governance energy. Suppose that for every nonterminal state $s_t$, the shielded action set $\mathcal{A}_{safe}(s_t)$ contains either a descent action satisfying $\mathbb{E}[V_K,(s_{t+1}) \mid s_t a] - V_K(s_t) \leq -\epsilon$, for some $\epsilon > 0$, or a terminal routing action $a_{term} \in \mathcal{A}_{term}(s_t)$ that routes the contested action to Assisted or Regulated termination. If Q-learning is used, we assume only shield-approved actions may execute: $a_t^{exec} \in \mathcal{A}_{safe}(s_t)$.

**Table 7. State Transition Paths**

| Path | Condition | LBCF interpretation |
|---|---|---|
| $S \to M \to A$ | Elevated but noncritical ambiguity | Internal recovery fails; contested action routes to Assisted validation |
| $S \to M \to Rt$ | Critical bag or threshold breach | Shield bypasses Assisted and routes to Regulated |
| $A \to Rt$ | Validator identifies missed critical context | External input changes predicate status; autonomy revoked |
| $S \to Rt$ | Privacy/cyber/safety predicate fails immediately | Direct hard halt before ordinary recovery |
| $A \to M \to S$ | Validator authorizes contextual update | $I_t \leftarrow I(X_0)'$; risk recomputed below $\tau_{safe}$ |

Then the LBCF process cannot continue indefinitely in autonomous execution of the contested action. It must either follow a bounded sequence of descent transitions or terminate in an Assisted or Regulated governance condition.

**Table 8: Scope for the LBCF convergence guarantee**

| Element | Assumed or defined | Not claimed |
|---|---|---|
| Neural monitoring | $D_t$, $B_{KL,t}$, and $R_t$ are measurable | Safety of the full activation space |
| Invariant reference | $I_t$ is available for comparison | Permanent model internalization of SME rules |
| Finite abstraction | Fixed $\phi: \mathbb{R}^3 \times \mathcal{G} \to \mathcal{S}_K$ | Lossless or universal abstraction |
| Q-learning | Q-values rank candidate actions | Unconstrained RL safety |
| Shield | Executed actions are filtered by $\mathcal{A}_{safe}(s_t)$ | External system compliance |
| Terminal routing | Assisted/Regulated routes stop contested autonomy | Validator correctness or availability |
| IDSS cases | Illustrate finite transitions | Clinical validation |
| HAI study | Tests finite governance simulation | Live ICS or attack-detection validation |

Proof sketch. Because $\mathcal{S}_K$ is finite and $V_K(s) \geq 0$, only finitely many strict descent steps of size at least $\epsilon$ can occur before $V_K$ reaches its lower bound or no descent action remains. The number of descent steps is bounded by

$$N \leq \left\lceil \frac{V_K(s_0) - V_{\min}}{\epsilon} \right\rceil, \text{where } V_{\min} = \min_{s \in \mathcal{S}_K} V_K(s).$$

If no descent action is available, the shield routes the contested action to Assisted or Regulated termination. If Q-learning is used, its preferred action cannot execute unless it is shield-approved. Therefore, unbounded autonomous continuation of the contested action is excluded under the finite abstraction. This theorem applies to $s_t$, not to raw neural activations, live clinical outcomes, or deployed ICS actuation. The next section, Section VII evaluates this finite shielded governance process using HAI-derived ICS time-series data. The simulation instantiates $D_t$, $B_{KL,t}$, and $R_t$, maps them into $s_t$, evaluates $V_K(s_t)$, filters candidate actions through $\mathcal{A}_{safe}(s_t)$, and reports governance outcomes. The empirical object is the finite SMARt/LBCF abstraction, not production MoDA internals, live human-in-the-loop performance, or PLC-level actuation.

## VII. HAI-BASED DISCRETE-EVENT GOVERNANCE SIMULATION

The simulation uses 86 HAI 22.04 process variables [50]-[51] across Boiler, Turbine, Water Treatment, and Virtual HIL domains, and evaluates 120,000 Monte Carlo iterations across three parameter-grouping strategies and four noise levels. The study is a discrete-event governance simulation, not a live deployment, attack-detection benchmark, or validation of a production MoDA transformer. HAI-derived process variables provide a cyber-physical operating context; controlled synthetic perturbations test whether the shielded SMARt/LBCF process maintains bounded convergence and prevents configured guard violations under fixed thresholds. The simulation asks a limited governance question: Given fixed SME-style invariants, threshold settings, and a finite K-bag abstraction, does the shielded SMARt/LBCF process converge or safely route contested actions without permitting configured guard violations? It does not evaluate attack detection, production MoDA internals, real-time human review, external-system availability, or PLC-level actuation. The simulation (configuration shown in Table 9) uses 86 HAI 22.04 process variables after excluding timestamp and label fields. The variables are drawn from four process domains: Boiler, Turbine, Water Treatment, and Virtual HIL simulators. These variables are used to instantiate process-structured governance signals rather than to train an attack classifier. Each run computes $\Psi_t = (D_t, B_{KL,t}, R_t)$, maps the tuple into $s_t = \phi(D_t, B_{KL,t}, R_t, g_t)$, evaluates $V_K(s_t)$, and filters governance actions through $\mathcal{A}_{safe}(s_t)$. If Q-learning is active, $Q(s_t, a_t)$ *will* rank candidate actions, but execution remains shielded:

$$a_t^{exec} \in \arg \max_{a \in \mathcal{A}_{safe}(s_t)} Q(s_t, a).$$

**Table 9. Simulation configuration**

| Component | Configuration |
|---|---|
| Dataset | HAI 22.04 ICS time-series records |
| Evaluation type | Discrete-event governance simulation |
| Process variables | 86 variables, excluding timestamp and label fields |
| Process domains | Boiler, Turbine, Water Treatment, Virtual HIL |
| Grouping strategies | Attack Susceptibility, Process Criticality, Sensor Type |
| Noise levels | 5%, 10%, 15%, 30% |
| Runs | 120,000 Monte Carlo iterations |
| Governance signals | $D_t$, $B_{KL,t}$, $R_t$ |
| Abstraction | Finite K-bag state space |
| Policy estimation | Shielded Q-learning or fixed shielded rules |
| Not evaluated | Attack detection, live MoDA, real-time HITL, PLC actuation |

Controlled Gaussian noise is injected at $5\%, 10\%, 15\%, 30\%$ to simulate sensor degradation and stochastic perturbation. This noise model does not fully represent coordinated adversarial spoofing or multi-stage ICS compromise. The simulation (Table 10) compares three parameter groupings: Attack Susceptibility, Process Criticality, and Sensor Type. The purpose is to test whether the finite governance process remains stable under different K-bag abstractions, not to prove that any grouping is universally optimal.

**Table 10. Attack susceptibility grouping**

| Group | Attack vector and definition | Vulnerability | Simulated noise range |
|---|---|---|---|
| P1 | Hard targets: core physical sensors, such as boiler pressure | Low | 2%–10% |
| P2 | Easy targets: actuator commands, such as valves or pumps | High | 5%–30% |
| P3 | Well-monitored targets: network pathways and logs | Moderate | 5%–20% |
| P4 | Control flags: mode switches and operational toggles | High-gravity | 5%–30% |
| P5 | Support systems: auxiliary or virtual infrastructure | Low | 5%–30% |

### *A. Simulation Results*

Across 120,000 simulated runs, as shown in Table 11, we observed 100.0% convergence, an average of 1.0 governance-level decision per run, mean energy descent around $-0.0068$, and zero configured guard violations across all three grouping strategies. Figure 4 shows how HAI-derived process variables are grouped into alternative K-bag abstractions, including Attack Susceptibility, Process Criticality, and Sensor Type. Figure 5 shows the Lyapunov energy trajectory, indicating that the finite governance state descends toward the configured convergence region under the tested simulation conditions. Figure 6 compares energy descent rates across grouping strategies and noise levels, showing that the observed descent remains consistent with the configured minimum descent requirement. Figure 7 shows the SMARt state-machine behavior with hysteresis-guarded transitions; under the selected thresholds, most simulated runs remained Stable, limited escalation occurred, and Regulated mode was not reached. These figures should be interpreted as evidence of finite-abstraction governance behavior under fixed thresholds and synthetic perturbations, not as live ICS validation or proof that Regulated intervention is unnecessary in deployment. These figures summarize the behavior of the finite SMARt/LBCF governance simulation.

Our results support a limited claim: under the tested finite abstraction, thresholds, and perturbation regimes, the shielded governance process converged and did not permit configured guard violations. The result should not be read as proof of general ICS safety, attack-detection capability, or continuous neural-state convergence. The reported one-step governance convergence reflects the simplified divergence model and fixed threshold schedule. The submitted paper already notes that real systems may require more steps because of physical inertia, latency, and operational timing, and that DNRS is positioned at a Level 2/3 supervisory tier rather than direct Level 1 PLC control. Regulated mode was not reached under the selected thresholds and perturbation regimes. This does not imply that Regulated intervention is unnecessary. In deployment, coordinated spoofing, missing telemetry, conflicting authoritative sources, privilege violations, or hard policy breaches may force Assisted or Regulated routing.

**Table 11. LBCF simulation results under tested thresholds**

| Parameter grouping | Trials | Convergence | Avg. governance decisions | Mean energy descent | Guard violations |
|---|---|---|---|---|---|
| Attack Susceptibility | 40,000 | 100.00% | 1.00 | $-0.00683$ | 0 |
| Process Criticality | 40,000 | 100.00% | 1.00 | $-0.00682$ | 0 |
| Sensor Type | 40,000 | 100.00% | 1.00 | $-0.00682$ | 0 |
| Aggregate | 120,000 | 100.00% | 1.00 | approx. $-0.0068$ | 0 |

### *B. Limitations*

Several limitations define the scope of the simulation. First, the divergence model is simplified and does not fully capture coupled cyber-physical dynamics. Second, HAI attack labels are not used to evaluate anomaly detection or attack classification; the study evaluates governance behavior under HAI-derived sensor conditions with synthetic perturbations. Third, thresholds and hyperparameters are fixed rather than adaptively tuned for each site. Fourth, Gaussian noise approximates sensor degradation but does not fully model coordinated adversarial spoofing or multi-stage ICS compromise. Fifth, the simulation does not evaluate production MoDA internals, live human-in-the-loop review, operator workload, SSO-protected external access, or PLC-level actuation. Sixth, zero guard violations are limited to the configured invariants, thresholds, and perturbation regimes. These limitations do not negate the simulation result. They define it. The study supports the internal consistency of shielded LBCF governance under a finite abstraction; it does not establish general deployment safety.

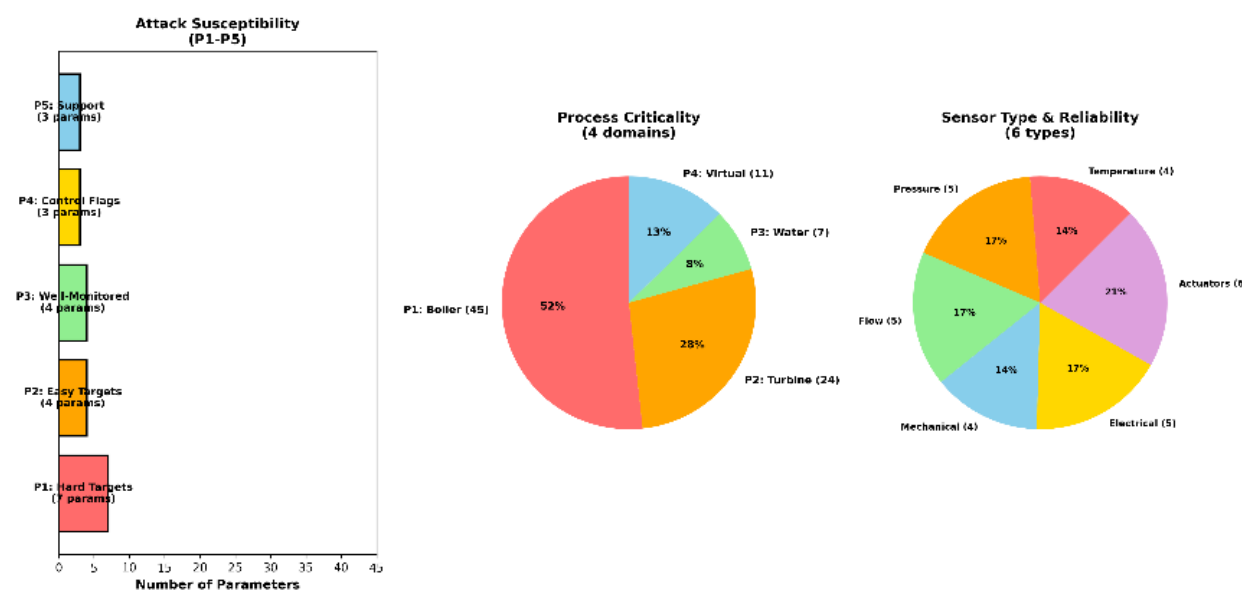

**Figure 4. Parameter grouping comparison under the finite K-bag abstraction. Compares three abstractions for organizing HAI-derived process variables and that equivalent convergence reflects the tested thresholds and perturbation model**

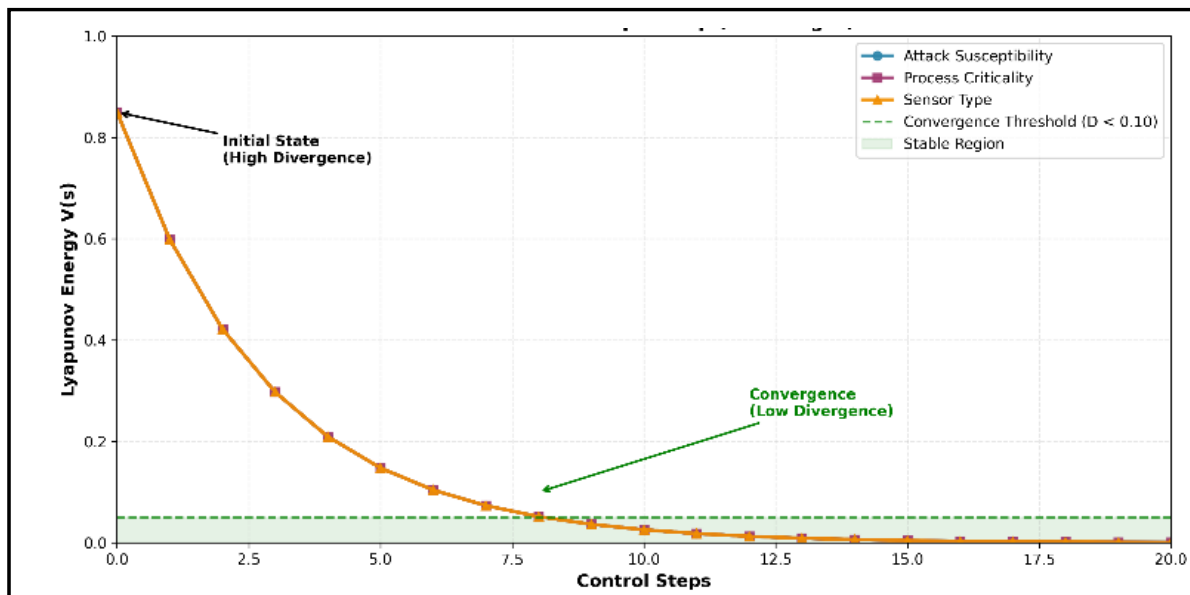

**Figure 5. Lyapunov energy trajectory under the finite LBCF simulation: Observed descent is consistent with the configured finite-state descent condition.**

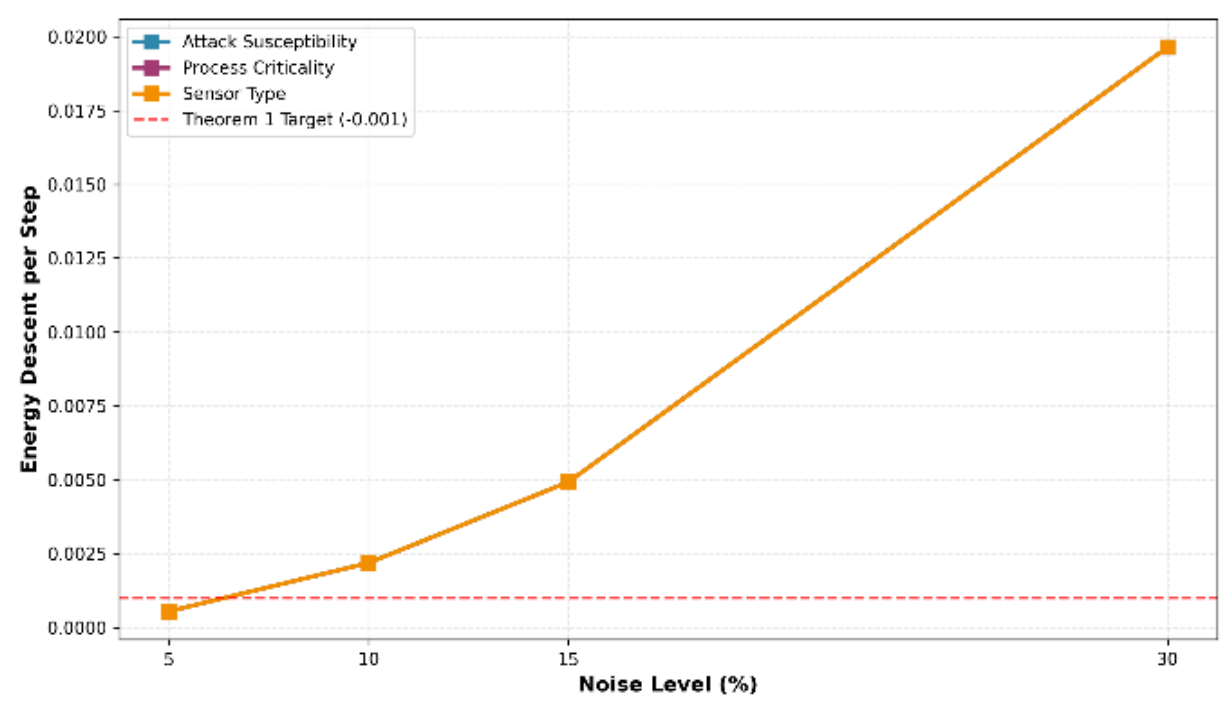

**Figure 6. Energy descent rate under controlled noise levels. Energy descent result applies to the tested synthetic perturbations.**

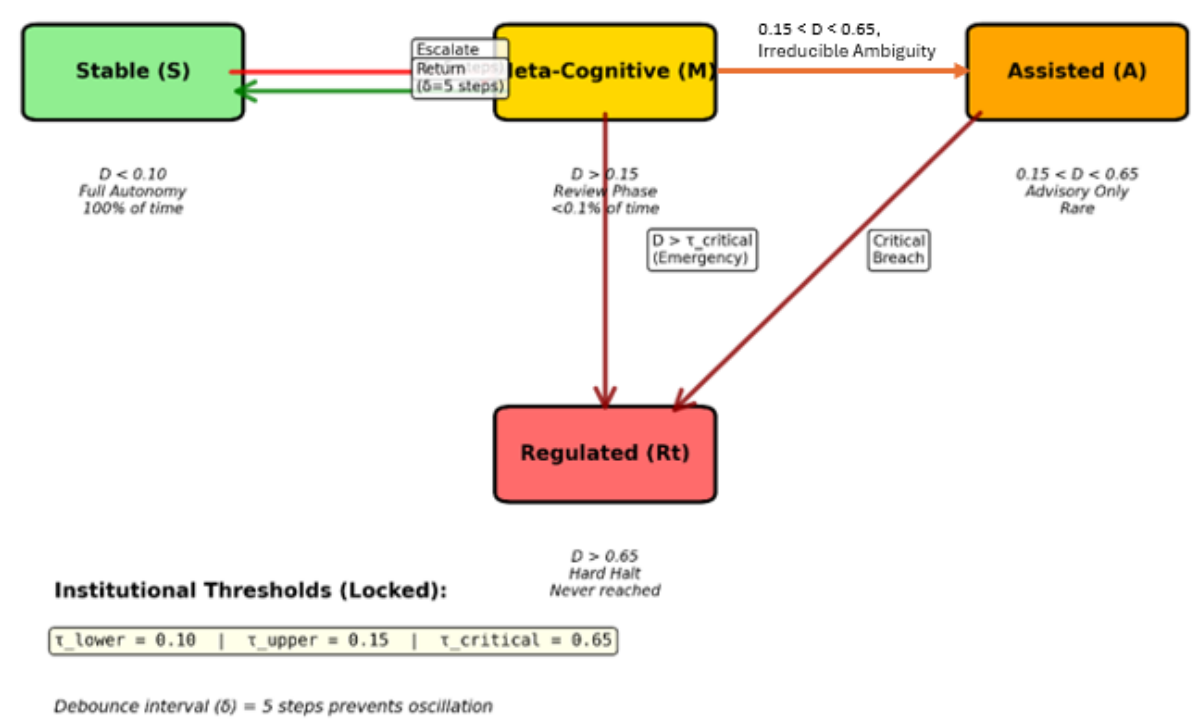

**Figure 7. SMARt state-machine governance with hysteresis-guarded transitions. Four states (Stable → Meta-Cognitive → Assisted → Regulated) Observed state distribution reflects selected thresholds, not proof that Regulated mode is unnecessary Regulated state was never reached.**

## VIII. CONCLUSIONS

DNRS offers a bounded governance architecture for agentic AI-enabled automation in safety-critical settings. Rather than treating model confidence as a binary permission signal, DNRS combines SME-derived invariant anchors, MoDA-style invariant access, V&V parameter surfacing, SMARtAutonomy state transitions, and the Lyapunov-Bounded Consensus Framework to translate continuous epistemic degradation into discrete autonomy-control decisions. The central claim is bounded: DNRS does not prove that the underlying neural model is intrinsically safe. It shows that, under a fixed finite abstraction and feasible shield assumptions, contested autonomous actions must either descend toward lower-risk governance states or route to Assisted or Regulated intervention. While the current DNRS architecture proposes MoDA-style internal access, harness-engineering principles [52] such as stateful cognitive offloading and structured action editing offer a path to decoupling the safety-governance framework from specific model architectures. The HAI-based study provides simulation-based evidence for this finite governance claim under fixed thresholds, synthetic perturbations, and K-bag abstraction assumptions. These results should be read as supervisory governance validation, not live ICS validation, attack-detection validation, or production MoDA validation. Future work will extend DNRS toward large-scale operational settings through Sparse Moment-Informed LBCF, federated multi-agent governance, broader empirical validation, and modular runtime interfaces. Sparse LBCF is especially important for defense-scale systems, where thousands of sensors, agents, documents, threat indicators, and mission constraints may interact, making full covariance tracking or exhaustive state enumeration infeasible. A companion line of work, Entropy Runtime, laying the next foundation by separating SMARt governance states from lower-level execution gears, utility-gated dispatch, fallback handling, and multi-agent runtime coordination, extending DNRS/LBCF-style governance toward distributed agentic AI systems.

### *Data Availability Statement*

The data supporting the findings of this study are derived from the publicly available HIL-based Augmented ICS Security

(HAI 22.04) dataset cited in References [51]. The simulation methodology, parameter groupings, invariant thresholds, noise levels, and evaluation metrics used in this study are described in the manuscript. No proprietary, private, or personally identifiable datasets were used. The study did not generate a new external dataset repository; derived simulation outputs and intermediate materials may be made available from the corresponding author upon reasonable request.

## IX. REFERENCE

[1] S. Gu et al., "A Review of Safe Reinforcement Learning: Methods, Theories, and Applications," *IEEE Transactions on Pattern Analysis and Machine Intelligence*, vol. 46, no. 12, pp. 11216-11235, Dec. 2024.

[2] J. Garcia and F. Fernandez, "A Comprehensive Survey on Safe Reinforcement Learning," *Journal of Machine Learning Research (JMLR)*, vol. 16, pp. 1437-1480, 2015.

[3] G. Dulac-Arnold, N. Levine, D. J. Mankowitz, et al., "Challenges of real-world reinforcement learning: definitions, benchmarks and analysis," *Machine Learning*, vol. 110, no. 9, pp. 2419–2468, Sept. 2021.

[4] Y. Chow, O. Nachum, E. Duenez-Guzman, A. Faust, and M. Ghavamzadeh, "A Lyapunov-Based Approach to Safe Reinforcement Learning," *Advances in Neural Information Processing Systems (NeurIPS)*, vol. 31, 2018.

[5] V. Borkar and R. Jain, "Risk-constrained Markov decision processes," *IEEE Transactions on Automatic Control*, vol. 59, no. 9, pp. 2574–2579, 2014.

[6] E. Altman, *Constrained Markov Decision Processes*. Boca Raton, FL, USA: Chapman & Hall/CRC, 1999.

[7] N. Jansen, B. Könighofer, S. Junges, A. Serban, and R. Bloem, "Safe reinforcement learning using probabilistic shields," in *Proceedings of the 31st International Conference on Concurrency Theory (CONCUR)*, vol. 171, pp. 3:1–3:16, 2020.

[8] D. Essen, L. Schaefer, and U. Topcu, "A Lyapunov Approach for Time-Bounded Reachability of CTMCs and CTMDPs," *IEEE Conference on Decision and Control (CDC)*, 2019.

[9] J. Yao, M. Han, and X. Yin, "Lyapunov-based distributed reinforcement learning control with stability guarantee," *Computers & Chemical Engineering*, vol. 195, 108979, 2025.

[10] M. Zanon, S. Gros, and M. Palladino, "Stability-Constrained Markov Decision Processes Using MPC," *Automatica*, 2022.

[11] N. Monir, O. Schön, and S. Soudjani, "Lyapunov-Based Policy Synthesis for Multi-Objective Interval MDPs," *IFAC-PapersOnLine*, vol. 58, issue 11, pp. 99–106, 2024.

[12] A. D. Ames, X. Xu, J. W. Grizzle, and P. Tabuada, "Control barrier function based quadratic programs for safety critical systems," *IEEE Transactions on Automatic Control*, vol. 62, no. 8, pp. 3861–3876, 2017.

[13] R. Cheng, G. Orosz, R. M. Murray, and J. W. Burdick, "End-to-end safe reinforcement learning through barrier functions for safety-critical continuous control tasks," in *Proceedings of the AAAI Conference on Artificial Intelligence*, vol. 33, no. 1, pp. 3387–3395, 2019.

[14] M. Alshiekh, R. Bloem, R. Ehlers, B. Könighofer, S. Niekum, and U. Topcu, "Safe reinforcement learning via shielding," in *Proceedings of the AAAI Conference on Artificial Intelligence*, vol. 32, no. 1, 2018.

[15] J. Kaplan, S. McCandlish, T. Henighan, et al., "Scaling laws for neural language models," *arXiv preprint arXiv:2001.08361*, 2020.

[16] H. Wang, S. Ma, L. Dong, et al., "DeepNet: Scaling Transformers to 1,000 Layers," *IEEE Transactions on Pattern Analysis and Machine Intelligence (PAMI)*, 2024.

[17] C. Chen and L. Wei, "Post-LayerNorm Is Back: Stable, ExpressivE, and Deep," *arXiv preprint arXiv:2601.19895*, 2026.

[18] K. He, X. Zhang, S. Ren, and J. Sun, "Deep residual learning for image recognition," in *Proceedings of the IEEE Conference on Computer Vision and Pattern Recognition (CVPR)*, 2016, pp. 770–778.

[19] M. Pagliardini, A. Mohtashami, F. Fleuret, and M. Jaggi, "Denseformer: Enhancing information flow in transformers via depth weighted averaging," *Advances in Neural Information Processing Systems (NeurIPS)*, 2024.

[20] Y. Bai, S. Kadavath, S. Kundu, et al., "Constitutional AI: Harmlessness from AI Feedback," *arXiv preprint arXiv:2212.08073*, 2022.

[21] W. Saunders, C. Snell, A. Askell, et al., "Self-Critique and Self-Reward: Training Language Models to Assess Their Own Generations," *arXiv preprint arXiv:2206.05802*, 2022.

[22] Y. Du, S. Li, A. Torralba, J. B. Tenenbaum, and I. Mordatch, "Improving factuality and reasoning in language models through multiagent debate," *Proceedings of the 41st International Conference on Machine Learning (ICML)*, 2024.

[23] G. Mialon, R. Dessì, M. Lomeli, et al., "Augmented Language Models: A Survey," *arXiv preprint arXiv:2302.07842*, 2023.

[24] N. Shinn, F. Cassano, A. Gopinath, et al., "Reflexion: Language Agents with Verbal Reinforcement Learning," *Advances in Neural Information Processing Systems (NeurIPS)*, 2023.

[25] S. Zhang, X. L. Li, Y. Shen, et al., "Safety Alignment for Multi-Agent Systems and LLM-Based Agents: A Survey," *arXiv preprint arXiv:2401.10605*, 2024.

[26] G. Huang, Z. Liu, L. Van Der Maaten, and K. Q. Weinberger, "Densely connected convolutional networks," in *Proceedings of the IEEE Conference on Computer Vision and Pattern Recognition (CVPR)*, 2017, pp. 4700–4708.

[27] A. Vaswani, N. Shazeer, N. Parmar, et al., "Attention is all you need," *Advances in Neural Information Processing Systems (NeurIPS)*, 2017.

[28] L. Zhu, Y. Fang, B. Liao, et al., "Mixture-of-Depths Attention," *arXiv preprint arXiv:2603.15619*, Mar. 2026.

[29] D. Raposo, S. Ritter, B. Richards, et al., "Mixture-of-Depths: Dynamically allocating compute in transformer-based language models," *arXiv preprint arXiv:2404.02258*, 2024.

[30] A. Gadhikar, S. K. Majumdar, N. Popp, et al., "Attention Is All You Need for Mixture-of-Depths Routing," *arXiv preprint arXiv:2412.20875*, 2024.

[31] M. Abdar, F. Pourpanah, S. Hussain, et al., "A review of uncertainty quantification in deep learning: Techniques, applications and challenges," *Information Fusion*, vol. 76, pp. 243-297, 2021.

[32] B. Wright, et al., "An Analytic Solution to Covariance Propagation in Neural Networks," *Proceedings of Machine Learning Research (PMLR)*, vol. 238, 2024.

[33] E. Daxberger, A. Kristiadi, A. Immer, et al., "Laplace Redux - Effortless Bayesian Deep Learning," *Advances in Neural Information Processing Systems*

*(NeurIPS)*, vol. 34, pp. 20089-20103, 2021.

[34] A. Le Folgoc, V. Baltatzis, S. Desai, et al., "Is MC-Dropout Calibrated for High-Dimensional Predictive Distributions?" *Advances in Neural Information Processing Systems (NeurIPS)*, vol. 34, 2021.

[35] D. A. Patterson and J. L. Hennessy, Computer Organization and Design: The Hardware/Software Interface. Morgan Kaufmann, 2013.

[36] M. V. Wilkes, "The best way to design an automatic calculating machine," in *Report of the Manchester University Computer Inaugural Conference*, 1951, pp. 16-18.

[37] E. M. Clarke, E. A. Emerson, and A. P. Sistla, "Automatic verification of finite-state concurrent systems using temporal logic specifications," *ACM Transactions on Programming Languages and Systems (TOPLAS)*, vol. 8, no. 2, pp. 244-263, 1986.

[38] M. Abadi, M. Budiu, U. Erlingsson, and J. Ligatti, "Control-flow integrity," in *Proceedings of the 12th ACM Conference on Computer and Communications Security (CCS)*, 2005, pp. 340-353.

[39] C. Zimmer, B. Bhat, F. Mueller, and S. Mohan, "Time-based intrusion detection in cyber-physical systems," in *Proceedings of the 1st ACM/IEEE International Conference on Cyber-Physical Systems (ICCPS)*, 2010, pp. 109–118.

[40] S. Malik and P. Subramanyan, "Specification and Modeling for Systems-on-Chip Security Verification," in *Proceedings of the 53rd Annual Design Automation Conference (DAC)*, 2016.

[41] J. Kong, O. Aciicmez, J.-P. Seifert, and H. Zhou, "Architecting against Software Cache-Based Side-Channel Attacks," *IEEE Transactions on Computers*, vol. 62, no. 7, pp. 1276–1288, 2013.

[42] A. Suraj, S. Ramaswamy, and K. S. Barber, "Behavior Specification of Sensible Agents," in *Workshop on Semiotic Modeling for Sensible Agents, Intelligent Systems: A Semiotic Perspective*, NIST, Gaithersburg, MD, Oct. 1996.

[43] A. Suraj, S. Ramaswamy, and K. S. Barber, "Extended StateCharts for the Modeling and Specification of Manufacturing Control Software Systems," *International Journal of Computer Integrated Manufacturing Systems*, vol. 10, no. 1-4, pp. 160-171, Jan. 1997.

[44] A. Suraj, S. Ramaswamy, and K. S. Barber, "An Integrated Approach to Design Reusability," in *First World Conference on Integrated Design and Process Technology*, Dec. 1995.

[45] K. S. Barber, T. H. Liu, A. Goel, and S. Ramaswamy, "A Flexible Reasoning Mechanism for Trade-off of System Versus Local Goals in Sensible Agents," in *Rensselaer's International Conference on Agile, Intelligent, and Computer-Integrated Manufacturing*, Oct. 1998.

[46] S. Ramaswamy, A. Suraj, and K. S. Barber, "An Approach for Monitoring and Control of Agent Based Systems," in *Proceedings of the IEEE International Conference on Robotics and Automation (ICRA)*, Albuquerque, NM, Apr. 1997.

[47] S. Ramaswamy, "A Petri Net Based Approach for Establishing Necessary Software Design and Testing Requirements," in *Proceedings of the IEEE International Conference on Systems, Man and Cybernetics*, Nashville, TN, Oct. 2000.

[48] S. Ramaswamy, "Intelligent Coordinating Entities Based Control Software Design," in *NSF Design and Manufacturing Grantees Conference*, Vancouver, Canada, Jan. 2000.

[49] S. Ramaswamy, "Intelligence as Managed Autonomy: Failure, Escalation, and Governance for Agentic AI Systems," *Journal of Intelligent & Robotic Systems*, to appear (accepted May 2026).

[50] H. K. Shin, W. Lee, J. Yun, and H. Kim, "HAI 1.0: HIL-based augmented ICS security dataset," *13th USENIX Workshop on Cyber Security Experimentation and Test (CSET 20)*, 2020.

[51] ICS Security Dataset Project, "HAI 22.04: HIL-based Augmented ICS Security Dataset," *GitHub repository*, 2022. [Online]. Available: https://github.com/icsdataset/hai

[52] P. Jiang, Z. Shi, K. Hong, et al., "Harness-1: Reinforcement Learning for Search Agents with State-Externalizing Harnesses", *arXiv preprint arXiv:2606.02373*, June 2026.